%% file: neurips_2025.tex
\documentclass{article}

\usepackage[preprint]{neurips_2025}

\usepackage[utf8]{inputenc} 
\usepackage[T1]{fontenc}    
\usepackage{hyperref}       
\usepackage{url}            
\usepackage{booktabs}       
\usepackage{amsfonts}       
\usepackage{amsmath}
\usepackage{nicefrac}       
\usepackage{microtype}      
\usepackage{xcolor}         

\usepackage{booktabs}
\usepackage{multirow} 
\usepackage{multicol} 
\usepackage{bbding}
\usepackage{enumitem} 

\usepackage{hyperref}
\usepackage{url}
\usepackage{subcaption} 
\usepackage{graphicx} 
\usepackage{diagbox} 
\usepackage{adjustbox} 
\usepackage{wrapfig}
\usepackage{makecell} 
\usepackage{titletoc}
\newcommand\dataset{MIRAGE}
\newcommand\method{Logos}

\title{MIRAGE: Assessing Hallucination in Multimodal Reasoning Chains of MLLM}

\author{
   Bowen Dong\textsuperscript{\rm 1,2} \, Minheng Ni\textsuperscript{\rm 1,2} \, Zitong Huang\textsuperscript{\rm 1} \, Guanglei Yang\textsuperscript{\rm 1}\, Wangmeng Zuo\textsuperscript{\rm 1}\, Lei Zhang\textsuperscript{\rm 2} \\
   \textsuperscript{\rm 1}Harbin Institute of Technology \quad \textsuperscript{\rm 2}The Hong Kong Polytechnic University\\
   \small{cndongsky@gmail.com \, kodenii@outlook.com \, cslzhang@comp.polyu.edu.hk \, wmzuo@hit.edu.cn}
}

\begin{document}

\maketitle

\begin{abstract}
Multimodal hallucination in multimodal large language models (MLLMs) restricts the correctness of MLLMs. 
However, multimodal hallucinations are multi-sourced and arise from diverse causes. 
Existing benchmarks fail to adequately distinguish between perception-induced hallucinations and reasoning-induced hallucinations. 
This failure constitutes a significant issue and hinders the diagnosis of multimodal reasoning failures within MLLMs. 
To address this, we propose the {\dataset} benchmark, which isolates reasoning hallucinations by constructing questions where input images are correctly perceived by MLLMs yet reasoning errors persist.
%
{\dataset} introduces multi-granular evaluation metrics: accuracy, factuality, and LLMs hallucination score for hallucination quantification.
%
%
%
Our analysis reveals that (1) the model scale, data scale, and training stages significantly affect the degree of logical, fabrication, and factual hallucinations; (2) current MLLMs show no effective improvement on spatial hallucinations caused by misinterpreted spatial relationships, indicating their limited visual reasoning capabilities; and (3) question types correlate with distinct hallucination patterns, highlighting targeted challenges and potential mitigation strategies.
To address these challenges, we propose {\method}, a method that combines curriculum reinforcement fine-tuning to encourage models to generate logic-consistent reasoning chains by stepwise reducing learning difficulty, and collaborative hint inference to reduce reasoning complexity. {\method} establishes a baseline on {\dataset}, and reduces the logical hallucinations in original base models.
{\dataset} will be publicly available.
\end{abstract}

\input{sections/introduction}
\input{sections/related_work}
\input{sections/benchmark}
\input{sections/method}
\input{sections/experiments}

\input{sections/conclusion}

\bibliography{neurips_2025}{}
\bibliographystyle{plain}








\input{sections/appendix}

\end{document}

%% file: sections/introduction.tex
\section{Introduction}\label{sec:introduction}

Multimodal large language models (MLLMs)~\cite{gpt-4o,gemini,grok3} achieve advancements in multimodal perception~\cite{llava,internvl,qwen2vl}, as evidenced by standard benchmarks~\cite{mathvision,mme,chartqa,scienceqa,okvqa}. Recent studies further enhance their reasoning capacities through post-training~\cite{qvq,virgo,o1,cot,vic}. However, two critical challenges remain, \emph{i.e.}, erroneous visual perception that fabricates non-existent content, and defective logical reasoning yielding inconsistent conclusions. These multi-source hallucinations (stemming from distinct perceptual and cognitive origins) fundamentally limit the practical utility.

To quantitatively measure hallucination in MLLMs, several multimodal benchmarks have been applied to detect and measure multimodal hallucination in object recognition~\cite{mme,pope,mmvp,hallusionbench,seedbench} or academic reasoning~\cite{mathvista} aspects. 
%
%
Existing benchmarks~\cite{mme,pope,mmvp,hallusionbench,seedbench,mathvista} attempt to measure hallucinations via object recognition or academic tasks. 
However, two critical gaps remain. 
%
First, current evaluations fail to distinguish between different types of hallucinations, \emph{i.e.}, perception-induced hallucinations caused by inaccurate visual understanding and reasoning-induced hallucinations stemming from logical flaws, making it difficult to pinpoint errors.
%
Second, most benchmarks focus on validating the content of answers or intermediate steps, while lacking fine-grained evaluation of the reasoning process in terms of perception and logic, thereby hindering the ability to trace error propagation patterns.
This absence of hierarchical analysis spanning answer-level outputs, step-level intermediate results, and thought-level reasoning logic prevents systematic diagnosis of reasoning failures. Addressing these gaps is essential for building trustworthy MLLMs.


\begin{figure}
\begin{center}
\includegraphics[width=0.8\textwidth]{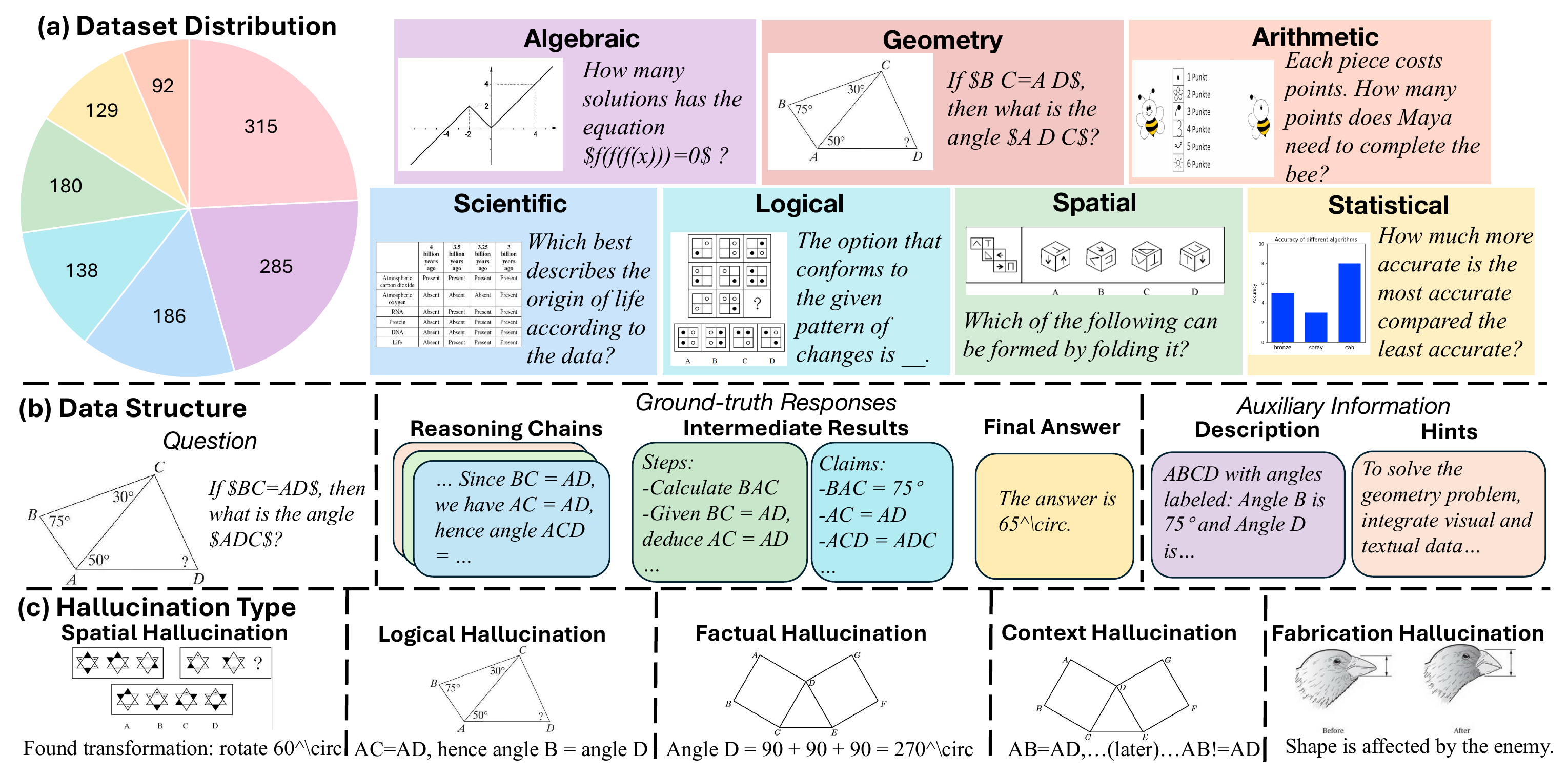}
\end{center}
\caption{Data distribution and structure of {\dataset} benchmark. (a) shows the classes and amounts of questions. (b) shows the data structure of each question in {\dataset}. And (c) shows the multimodal reasoning hallucination types we explored.
}
\label{fig:dataset_info}
\end{figure}

To address these challenges, we propose {\dataset}, a diagnostic benchmark specifically designed to isolate reasoning-induced hallucinations in MLLMs. As shown in Fig.~\ref{fig:dataset_info}, {\dataset} contains 1,329 questions where MLLMs demonstrate accurate visual perception but exhibit defective reasoning. Each question provides three-tier annotations: final answers, intermediate reasoning steps and claims, and ground-truth reasoning chains, enabling precise tracking of hallucination propagation in multimodal reasoning processes.
%
To comprehensively assess reasoning hallucination in MLLMs, {\dataset} proposes three evaluation metrics, \emph{i.e.}, accuracy assessment measuring overall answer correctness, factuality assessment verifying the correctness in intermediate steps and claims, and LLMs Hallucination Score to assess hallucination from the whole reasoning chain level. 
By evaluating MLLMs on {\dataset} from different levels, we aim to answer three critical research questions. 
%
First, how reasoning hallucinations compromise MLLM robustness and correlate with answer accuracy. 
Second, whether specific question types induce distinct hallucination patterns unique to multimodal reasoning. 
Finally, the efficacy of current mitigation methods against reasoning-specific hallucinations. 
%


We conduct extensive experiments on {\dataset}, leading to several key insights. First, the model scale, data scale, and training stages of MLLMs significantly influence the severity of logical, fabrication, and factual hallucinations. Second, these factors offer limited improvements in addressing spatial hallucinations, which are primarily caused by misinterpretations of spatial relationships—highlighting the limited visual reasoning capabilities of current MLLMs and their inability to benefit from straightforward scaling. Third, we observe strong correlations between question types and specific patterns of reasoning hallucination, underscoring critical challenges and suggesting targeted mitigation strategies. These findings offer valuable guidance for the future development of more reliable and reasoning-aware MLLMs.

Building on the insight that increasing the probability of logic-consistent reasoning chains reduces specific logical hallucinations, we propose {\method}, which integrates curriculum reinforcement fine-tuning (CRFT) for training and collaborative hint inference for testing. During training, CRFT with online reward filtration (ORF) gradually increases question difficulty while dynamically selecting high-reward samples, guiding the model toward accurate and logic-consistent reasoning. At the testing stage, collaborative hint inference provides topic- and question-specific hints from LLMs, reducing reasoning complexity for optimized models.
Experiments demonstrate that {\method} significantly reduces reasoning hallucinations and achieves strong performance on both {\dataset} and standard benchmarks~\cite{mathvista}.
%
%
%
In conclusion, the contribution of this paper can be summarized as follows:

\begin{itemize}
    \item We propose {\dataset}, the first benchmark for evaluating multimodal reasoning hallucinations in MLLMs. It isolates reasoning hallucinations with tasks where inputs are correctly perceived but reasoning errors persist, and introduces multi-level metrics for comprehensive assessment: accuracy, factuality, and LLMs hallucination score.
    \item Our findings reveal that the model scale, data scale, and training stages of MLLMs, and highlight critical challenges and mitigation for specific hallucination types. These findings will provide insights for future MLLM development.
    \item We propose {\method}, a baseline method of {\dataset} to encourage model for logic-consistent reasoning via curriculum reinforcement fine-tuning and collaborative hint inference. {\method} reduces the multimodal logical hallucination and improves the answer accuracy. 
\end{itemize}

\begin{table}[ht!]
    \centering
    \caption{
    Comparison of {\dataset} with existing benchmarks.
    ``MCQ'' means multiple-choice questions, ``A'' means answers, ``D'' means multimodal input descriptions, `R'' means full reasoning chains, and ``S'' means intermediate results. $^*$ means multiple reasoning chains. {\dataset} offers superior annotation coverage and assessment capabilities in reasoning hallucination assessment.
    }
\setlength{\tabcolsep}{1.3pt} %
    \renewcommand{\arraystretch}{2.5}%
    { \fontsize{8.3}{3}\selectfont{
\begin{tabular}{l|cccc|cc|c}
   \toprule
   \multirow{2}{*}{\bf{Benchmarks}} & \multicolumn{4}{c|}{\bf{Dataset Properties}} & \multicolumn{2}{c|}{\bf{Hallucination Assessment}} &  \multirow{2}{*}{\bf{Usage}}  \\ \cline{2-7}
   & {Taxonomy} & {Scale} & {Annotation} &{Intermediate} & {Steps} & {Chains}  \\
   \hline
   POPE~\cite{pope} & Object & 18K & A  & - & \XSolidBrush & \XSolidBrush & Object Hallu \\
   MMVP~\cite{mmvp} & Object & 300 & A & - & \XSolidBrush & \XSolidBrush & MCQ  \\
   HallusionBench~\cite{hallusionbench} & Object & 1,129 & A  & - & \XSolidBrush & \XSolidBrush & Illusion  \\
   MME~\cite{mme} & Various & 2,374 & A  & - & \XSolidBrush & \XSolidBrush & General VQA   \\
   SEEDBench~\cite{seedbench} & Obj+Act & 19K & A  & - & \XSolidBrush & \XSolidBrush & MCQ  \\
   MathVista~\cite{mathvista} & Math & 1,000 & A  & - & \XSolidBrush & \XSolidBrush & Math Reasoning \\ 
   OmniBench~\cite{omnibench} & Various & 1,142 & A/D  & - & \XSolidBrush & \XSolidBrush & MCQ \\
   MME-CoT~\cite{mme-cot} & {Various} & 1,130 & {A/D/S}  & Steps & \Checkmark & \XSolidBrush & General CoT \\
    \hline
    \bf{MIRAGE (Ours)} & \textbf{Various} & 1,329 & \textbf{A/D/R$^{*}$/S}  & \textbf{Steps+Hints}  & \Checkmark & \Checkmark & \textbf{Reason Hallu}  \\
   \bottomrule
\end{tabular}
}
}
\label{table:dataset_comp}
\end{table}

%% file: sections/related_work.tex
\section{Related Work}
\label{sec:related_work}
\noindent\textbf{Reasoning Multimodal Large Language Models. }\label{sec:related_cot}
%
Reasoning MLLMs can be roughly divided into three groups. First is the prompt-based reasoning method~\cite{cot,mllmcot,compositional} to guide MLLMs by in-context learning~\cite{incontext}. 
Second is the plan-based method~\cite{AGoT,BDoG,llamaberry}, which uses searching methods~\cite{mcts,prm} to explore optimal reasoning chains. 
And last is learning-based method by supervised fine-tuning (SFT)~\cite{r1-onevision} or reinforcement learning (RL)~\cite{deepseekmath,visual-rft,mm-eureka}. 
%
RL methods generalize better by optimizing with their high-reward predictions instead of fixed ground-truths. Hence, we build {\method} on RL for hallucination mitigation, uniquely focusing on dynamic training difficulty adjustment. 

\noindent\textbf{Multimodal Hallucination Evaluation. }\label{sec:related_hallu_eval}
Existing MLLMs still suffer from multimodal hallucination, where generated text either contradicts the visual input or deviates from correct logical reasoning.
To assess the hallucination and its effect in MLLMs, recent works measure the accuracy degradation among object perception~\cite{pope}, illusion~\cite{hallusionbench,autohallusion}, mathematic~\cite{mathvista,mathvision,mathverse}, IQ test~\cite{mmiq,puzzlevqa,algopuzzlevqa}, and general multimodal abilities~\cite{seedbench,mme}.
%
While existing benchmarks have advanced multimodal evaluation, they often conflate perception-induced hallucinations with reasoning-induced ones, making it challenging to diagnose reasoning failures.
In contrast, {\dataset} focuses on reasoning hallucinations by isolating reasoning errors from correctly perceived inputs, providing multi-level metrics for assessment.

\noindent\textbf{Multimodal Hallucination Mitigation. }\label{sec:related_hallu_mitigation}
Hallucination evaluations provide critical insights for future mitigation strategies, which broadly fall into training-free and training-based approaches.
First is training-free methods~\cite{vic,selfcorrection,react}. They use prompting~\cite{SoM-LLaVA}, question decomposition~\cite{vic}, reasoning chain reflection~\cite{selfcorrection,ji-etal-2023-towards}, and ensembling~\cite{woo2024ritual} to mitigate perception-induced hallucination.
Second is training-based mitigation methods~\cite{ursa,r1-onevision,virgo,yang2025mitigating,zhang2024reflective,yu2024hallucidoctor}. These methods mitigate hallucination by data curation~\cite{yu2024hallucidoctor,lrv-instruction}, SFT~\cite{sarkarmitigating,ben2023mocha,jiang2024hallucination,chen2023mitigating}, and RL~\cite{ha-dpo,zhou2024aligning,yu2024rlhf}.
%
Previous hallucination mitigation methods mainly target perception errors in MLLMs, with only a few early efforts addressing multimodal reasoning hallucinations~\cite{vic,ursa}.
Our proposed {\method} focuses on reasoning hallucinations and shows that combining CRFT with CHI effectively reduces hallucinations beyond prior methods.

%% file: sections/benchmark.tex
\section{MIRAGE Dataset}
\label{sec:benchmark}


\subsection{Data Construction}\label{sec:benchmark_construction}
\begin{figure}
\begin{center}
\includegraphics[width=0.9\textwidth]{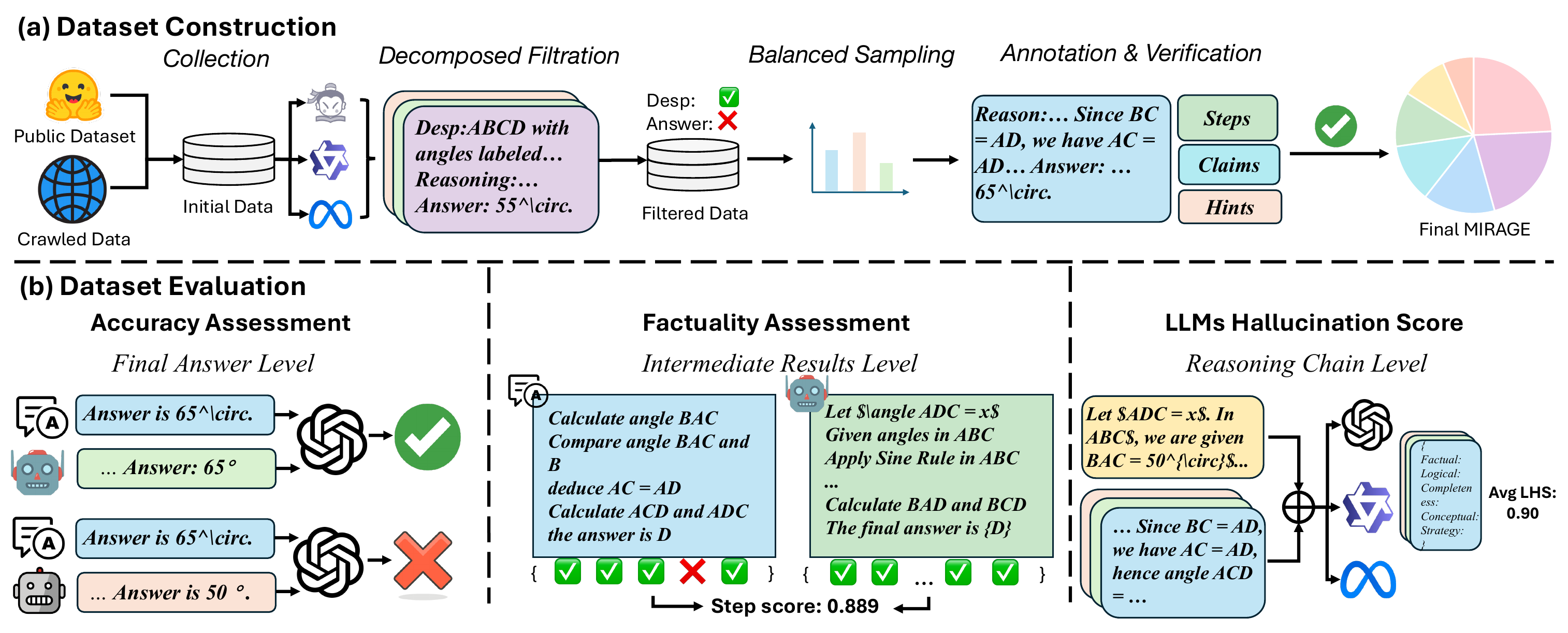}
\end{center}
\caption{The construction and evaluation of {\dataset}. (a) shows the construction of {\dataset}.
And (b) shows multi-granular evaluation metrics: accuracy, factuality, and LLMs hallucination score.
}
\label{fig:dataset_construct_eval}
\end{figure}

To evaluate reasoning hallucinations, as in Table~\ref{table:dataset_comp}, we present {\dataset}, emphasizing tasks with accurate perception but challenging reasoning. {\dataset} offers multi-level annotations and rich auxiliary data for error diagnosis. As in Fig.~\ref{fig:dataset_construct_eval}, the construction involves data collection and curation.




\noindent\textbf{Data Collection. }
To systematically evaluate multimodal reasoning capacities across diverse cognitive dimensions, the {\dataset} is constructed through rigorous selection of seven distinct taxonomies, including geometry, algebraic, arithmetic, scientific, spatial reasoning, and statistical reasoning. 
%
%
%
%
Based on above taxonomies, we collect the original benchmark data from both publicly available datasets and questions from Internet. 
Finally, the size of original dataset is roughly 18K.

\noindent\textbf{Data Curation. }
To ensure {\dataset} isolates reasoning hallucinations and comprehensively evaluate each topic, we apply a two-step curation process, \emph{i.e.}, difficulty curation and balance curation.
For difficulty curation, we use three open-source MLLMs~\cite{qwen25vl, internvl, llama3} to generate image descriptions, retaining only questions where these descriptions are consistently accurate (verified by a secondary LLM~\cite{deepseek-v3}) but lead to frequent reasoning errors, aligning with our benchmark focus.
For balance curation, we sample the resulting data to ensure a balanced distribution~\cite{lpt,kangdecoupling,li2022long} across seven topics, maintaining a small imbalanced rate, and resulting in a final dataset of 1,329 questions.

\subsection{Data Annotation and Verification}\label{sec:benchmark_annotation}
\noindent\textbf{Reasoning Chain Annotation. }
To address the lack of ground-truth reasoning chains, we propose a cost-effective automated annotation framework that optimizes both computational efficiency and output quality. Our approach follows a two-stage refinement process. Firstly, we generate initial reasoning chains using the lightweight O3-mini~\cite{o1}. And then we refine these chains with a strong LLM~\cite{deepsek-r1}, guided by known ground-truths. We will discuss the annotation cost in appendix.

\noindent\textbf{Collaborative Verification. }
Next, we conduct annotation verification to ensure the correctness.
To improve both verification speed and accuracy, we introduce a human-AI collaborative verification framework. Specifically, each question is independently assessed by a human expert and an MLLM~\cite{grok3} for potential hallucinations in the reasoning chain. If both assessments are accurate, the reasoning chain is retained as the ground-truth. In cases of discrepancies, the human expert either guides the MLLM to correct the reasoning chains or manually provides reasoning steps if the MLLM remains inaccurate. Finally, all newly annotated chains undergo cross-checking by other experts. 

\noindent\textbf{Step and Claim Extraction. }
Finally, given the verified reasoning chain and final answer for each question, we use a state-of-the-art LLM~\cite{deepseek-v3} to extract critical intermediate steps and claims via in-context learning. Specifically, for each ground-truth reasoning chain $\hat{\mathbf{y}}$, we use hand-crafted few-shot prompts to guide the LLM in selecting $\hat{K}_s$ representative reasoning steps $\hat{\mathbb{S}} = {\hat{\mathbf{s}}1,...,\hat{\mathbf{s}}_{K_s}}$ and $\hat{K}_c$ intermediate claims $\hat{\mathbb{C}} = {\hat{\mathbf{c}}1,...,\hat{\mathbf{c}}_{K_c}}$. To ensure reliability, we limit $1 \leq K_s \leq 10$ and $1 \leq K_c \leq 10$, preventing over-detailed and unreliable outputs. The final intermediate steps and claims are then parsed using regular expressions for fine-grained reasoning hallucination evaluation.

\noindent\textbf{Auxiliary Information Annotation.} {\dataset} also uses MLLM~\cite{gpt-4o} to annotate image descriptions and hints. This information is verified by experts and can help researchers to diagnose hallucinations.

\section{{\dataset} Benchmark Evaluation}\label{sec:evaluation}

\subsection{Accuracy Assessment}
The accuracy is a fundamental metric 
since incorrect \textbf{final answers} often indicate reasoning chain hallucinations~\cite{vic,receval}. To accommodate different question types, {\dataset} parses the final predictions and matches parsed answers $\mathbf{A}_{\text{pred}}$ with ground-truths $\mathbf{A}_{\text{gt}}$ for both multiple-choice and deterministic free-form answers. For questions with approximate answers (\emph{e.g.}, statistical questions without precise annotations on charts), {\dataset} calculates the relative error between predictions and ground-truths, considering answers correct if the error falls below a threshold.

\subsection{Factuality Assessment}\label{sec:evaluation_factuality}
\noindent\textbf{Step and Claim Factuality Evaluation. }
For each predicted reasoning chain $y$, {\dataset} follows the extraction pipeline in Sec.~\ref{sec:benchmark_annotation} to extract \textbf{intermediate} steps $\mathbb{S} = \{\mathbf{s}_{1},...,\mathbf{s}_{K_s}\}$ and claims $\mathbb{C} = \{\mathbf{c}_{1},...,\mathbf{c}_{K_c}\}$. With given corresponding ground-truth intermidiate steps $\hat{\mathbb{S}}$ and claims $\hat{\mathbb{C}}$, we utilize an LLM~\cite{deepseek-v3} and use $\{\mathbb{S}, \hat{\mathbb{S}}\}$ as input, and guide the LLM to detect whether a predicted step $\mathbf{s}_{i}$ is covered in $\hat{\mathbb{S}}$ and whether a ground-truth step $\hat{\mathbf{s}}_{i}$ is stated in $\mathbb{S}$, and then predicts the binary matching results $\mathbf{M}_{s,\text{pred}}$ and $\mathbf{M}_{s,\text{gt}}$. 
By this formulation, {\dataset} can efficiently match free-form steps and claims for flexible factuality evalution. 
Finally, we calculate the step factuality score $F_{\text{step}}$ by:
\begin{equation}
    F_{\text{step}} = \frac{2 \times \text{Precision}_s \times \text{Recall}_s}{\text{Precision}_s + \text{Recall}_s},
\end{equation}
where $\text{Precision} = \frac{|\mathbf{M}_{s,\text{pred}}=1|}{|\mathbf{M}_{s,\text{pred}}|}$ means correctly answered steps and $\text{Recall} = \frac{|\mathbf{M}_{s,\text{gt}}=1|}{|\mathbf{M}_{s,\text{gt}}|}$ means correctly matched ground-truth steps. Similarly, the claim factuality score $F_{\text{claim}}$ is defined by:
\begin{equation}
    F_{\text{claim}} = \frac{2 \times \text{Precision}_c \times \text{Recall}_c}{\text{Precision}_c + \text{Recall}_c}.
\end{equation}
\noindent\textbf{Hallucination Type Detection. }
Besides, to qualitatively assess which kind of reasoning hallucination does an MLLM suffer in a specific question, we follow LLM-as-a-Judge~\cite{fu-etal-2024-gptscore,gu2024survey} and introduce an LLM-based hallucination detector. Specifically, rather than compare two plain reasoning chains directly, {\dataset} detects the reasoning hallucination by examining extracted intermediate steps $\{\mathbb{S}, \hat{\mathbb{S}}\}$ and claims $\{\mathbb{C}, \hat{\mathbb{C}}\}$, then predict the hallucination detection results by in-context learning.

\subsection{LLMs Hallucination Score (LHS) Assessment}

Finally, we also assess the hallucination from \textbf{the whole reasoning chains}. 
While entropy-based uncertainty estimation methods~\cite{uncertainty_nature,vl-uncertainty,zhou2024relying,tomani2024uncertainty} can identify unreliable reasoning or information-deficient chains, they still face two limitations.
First is token-level likelihood dependencies. Existing methods rely on token-level likelihood to quantify uncertainty, but it is inaccessible in black-box MLLMs~\cite{gemini,gpt-4o,grok3}.
And second is high computational cost. Accurate uncertainty assessment typically requires sampling numerous responses per query, escalating evaluation overhead.
Therefore, inspired by LLM as judges~\cite{li2024salad,fu-etal-2024-gptscore,gu2024survey,li2024llms}, we propose LLMs Hallucination Score (LHS) to simulate uncertainty estimation via multi-LLMs and multi-reference ensemble. Specifically, 
we first define multi-dimension scoring rules to measure the hallucination in the whole reasoning chain rather than extracted steps, including factual accuracy, logical consistency, reasoning completeness, conceptual accuracy, and strategy appropriateness. Above dimensions can be used to simulate the uncertainty in responses and formulate the scoring template $\mathbf{h}_{\text{score}}$. Our aim is to predict LHS by $M$ (\emph{e.g.}, 3) LLM judgers. 
To improve the confidence of LHS, {\dataset} leverage an LLM~\cite{deepseek-v3} to rewrite $\hat{\mathbf{r}}$ by $N-1$ variants, thus formulating $N=3$ reference chains $\{\mathbf{r}_{\text{ref}}^{1},...,\mathbf{r}_{\text{ref}}^{N}\}$
Then, for each response $\mathbf{r}_{\text{ref}}^{i}$ with corresponding ground-truth $\hat{\mathbf{r}}$, both responses are integrated into the template $\mathbf{h}_{\text{score}}$ and then generate the judgement scores $\{s_1^{i,j},...,s_5^{i,j}\}$ by each $j$-th LLM judger. 
Finally, the LHS $\bar{s}$ of response $r$ is:
\begin{equation}
    \bar{s} = \frac{1}{M}\sum_{j=1}^{M}\frac{1}{N}\sum_{i=1}^{N}\text{mean}(\{s_1^{i,j},...,s_5^{i,j}\}).
\end{equation}
 By accumulating responses in {\dataset}, one can calcuate the mean and standard deviation of $\bar{s}$ for a specific reasoning MLLM.
Generally, lower mean indicates higher uncertainty (\emph{i.e.}, hallucination rate), and lower standard deviation means higher confidence of the LHS.
%
We further conduct consistency checks on LHS using human evaluators. We randomly sample 100 responses from Gemini-2-flash and Qwen2.5-VL-7B, comparing the human evaluation from three experts. The average difference rate is 7.5\%, showing the reliability of LHS for measuring reasoning hallucinations.

%% file: sections/method.tex
\section{{\method}: A Baseline Method of {\dataset}}
\label{sec:method}
\subsection{Revisit Multimodal Reinforcement Fine-Tuning}
As shown in Sec.~\ref{sec:exp_analysis}, reasoning chains with correct answers generally have lower hallucination rates (\emph{e.g.}, logical hallucination). This suggests that reducing hallucinations in MLLMs can be approached by increasing the generation probability of logic-consistent and correct reasoning chains, aligning inherently with Group Relative Policy Optimization (GRPO).
To address this, we propose the baseline method {\method} for {\dataset}, leveraging GRPO to optimize MLLM (the policy model $\pi$ with parameter $\theta$).
%
Specifically, we leverage in-context learning~\cite{cot} to guide $\pi$ to generate formatted response with ``\texttt{<think>...</think>}'' and ``\texttt{<answer>...</answer>}'' blocks, where the former contains the reasoning chain and the latter includes the final answer. 
Rather than using a separate value model to calculate advantages of responses, GRPO directly samples $G$ different responses $\{\mathbf{r}_{1},...,\mathbf{r}_{G}\}$ with given multimodal question $\mathbf{x}$.
To measure the relative advantages $\{A_{1},...,A_{G}\}$, we define the reward function $\mathcal{R}$ as format reward $\mathcal{R}_{\text{fmt}}$ and accuracy reward $\mathcal{R}_{\text{acc}}$, where the former is a binary function to judge whether the $i$-th response $\mathbf{r}_{i}$ follows response format, and the latter is a binary function to judge the correctness of final answer. Then the reward of $\mathbf{r}_{i}$ is defined by $r_i = \mathcal{R}_{\text{fmt}}(\mathbf{r}_{i}) + \mathcal{R}_{\text{acc}}(\mathbf{r}_{i})$. And the advantage $A_{i}$ is:
\begin{equation}
    A_{i} = \frac{r_i - \text{mean}(\{r_1,...,r_G\})}{\text{std}(\{r_1,...,r_G\})}.
\end{equation}
Finally, we optimize $\pi$ via minimizing GRPO loss $\mathcal{L}_{\text{GRPO}}$ with above advantages as follows:
\begin{equation}
\begin{aligned}
  \mathcal{L}_{\text{GRPO}} = -\mathbb{E}_{\{\mathbf{r}_{i}\}_{1}^{G} \sim \pi(\mathbf{x})}& \frac{1}{G} \sum_{i=1}^{G} \frac{1}{|\mathbf{r}_{i}|} \sum_{t=1}^{|\mathbf{r}_{i}|}  \min \left( \tilde{r}_{i,t}(\theta)A_{i}, \text{clip}(\tilde{r}_{i,t}(\theta), 1-\epsilon, 1+\epsilon)A_{i} \right) ,\\
  & \text{where } \tilde{r}_{i,t} = \frac{\pi(\mathbf{r}_{i,t}|\mathbf{x},\mathbf{r}_{i,1:t-1})}{\pi_{\text{old}}(\mathbf{r}_{i,t}|\mathbf{x},\mathbf{r}_{i,1:t-1})}.
\end{aligned}
\end{equation}
\begin{figure}
\begin{center}
\includegraphics[width=0.99\textwidth]{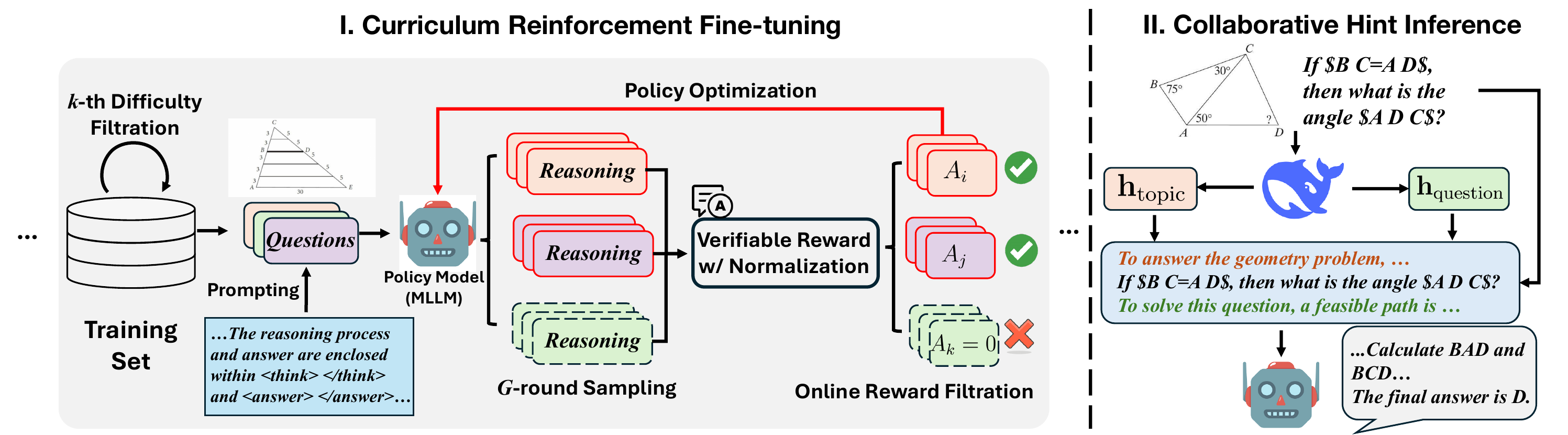}
\end{center}
\caption{
Overview of our baseline {\method}. During training, {\method} adopts curriculum reinforcement fine-tuning (CRFT) with online reward filtration (ORF) to progressively increase data difficulty and filter low-impact samples. 
During testing, {\method} introduces Collaborative Hint Inference, leveraging LLM-guided hints to simplify the reasoning process. Logos effectively reduces logical hallucination.
}
\label{fig:method}
\end{figure}
We remove the KL-divergence term since reasoning models have a non-negligible distribution gap with base models~\cite{lmm-r1}. We will investigate the effect in the App.~\ref{sec:app_more_analysis}.


\subsection{Curriculum Reinforcement Fine-Tuning}
Note that if responses from a specific training sample are all correct or all incorrect, the advantage of each response is 0, which is harmful for GRPO optimization. To reduce the difficulty and improve the training efficiency, we propose curriculum reinforcement fine-tuning (CRFT). 
Specifically, before the optimization, we first leverage $\pi$ to conduct $G$-round sampling, and calculate average accuracy reward $\bar{r}_{\text{acc}}=\text{mean}(\{\mathcal{R}(\mathbf{r}_1),...,\mathcal{R}(\mathbf{r}_G)\})$. During the first stage, we keep questions with $\bar{r}_{\text{acc}}>0$ to ensure that $\pi$ can sample at least one reasoning chain with correct answer and logic-consistent reasoning during training, thus making the advantages non-zero for smooth optimization. Then, during each $k$-round ($k>1$) curriculum training, we repeat $ G$-round sampling and keep questions with $\bar{r}_{\text{acc}}<0.5$ to ensure $\pi$ can face more difficult questions during further CRFT. Our experimental results in Sec .~\ref {sec:exp_ablation} illustrate the efficiency of CRFT. 

\begin{table}[ht]
    \centering
    \caption{Comparison of recent state-of-the-art MLLMs on {\dataset}. Best results are \textbf{bolded}. }
\setlength{\tabcolsep}{6pt} %
    \renewcommand{\arraystretch}{3.0}%
    { \fontsize{8.3}{3}\selectfont{
\begin{tabular}{lc|c|cc|c}
   \toprule
   \multirow{2}{*}{\bf Model}  & \multirow{2}{*}{\bf Type} & \multirow{2}{*}{\bf{Accuracy} $\uparrow$} & \multicolumn{2}{c|}{\bf{Factuality}} &  \multirow{2}{*}{\bf{LHS} $\uparrow$}  \\ \cline{4-5}
    &&&{\bf $F_{\text{step}}$ $\uparrow$} & {\bf $F_{\text{claim}}$ $\uparrow$}   \\
   \midrule
   
   
   \multicolumn{6}{c}{\emph{Black-Box MLLMs}} \\
    \hline
   Gemini-2-Flash-Thinking~\cite{gemini}  & Reasoning  & 47.6 & \bf 51.5 & \bf 50.7 & \bf 0.7517$\pm$0.0168  \\
   O1~\cite{o1}  & Reasoning & \bf 49.7 & 41.3 & 42.7 & 0.6193$\pm$0.0091  \\
   Gemini-2-Flash~\cite{gemini}  & General & 44.1 & 47.8 & 47.4 & 0.6882$\pm$0.0496   \\
   GPT-4o~\cite{gpt-4o}  & General  & 35.0  & 39.2 & 40.6 & 0.6332$\pm$0.0111 \\
   \hline
   \multicolumn{6}{c}{\emph{Open-sourced $\sim$72B MLLMs}} \\
    \hline
    {Qwen2.5-VL-72B-Instruct}~\cite{qwen25vl}  & General  & \bf 38.8 & \bf 47.4  & \bf 44.6 & \bf 0.7223$\pm$0.0339\\
    {InternVL-2.5-78B}~\cite{internvl}  & General  & 29.6 & 39.0 & 36.6 & 0.6377$\pm$0.0325\\
    {Qwen2-VL-72B-Instruct}~\cite{qwen2vl}  & General  & 24.5 & 29.7 & 26.2 & 0.4928$\pm$0.0332\\
    {QvQ-72B-Preview}~\cite{qvq}& Reasoning  & 31.0 & 46.1 & 45.3  & 0.5717$\pm$0.0597\\
    {Virgo-72B}~\cite{virgo} & Reasoning  & 37.4 & 47.1 & 45.0 & 0.6328$\pm$0.0251\\
   \hline
   \multicolumn{6}{c}{\emph{Open-sourced $\sim$7B MLLMs}} \\
    \hline
    {Qwen2.5-VL-7B-Instruct}~\cite{qwen25vl} & General  & 28.8 & 34.7 & 31.7 & 0.5996$\pm$0.0123\\
    {Qwen2-VL-7B-Instruct}~\cite{qwen25vl} & General  & 19.5 & 21.9 & 18.6 & 0.3633$\pm$0.0106\\
    {Qwen2.5-VL-7B-Instruct}+VIC~\cite{vic} & Reasoning  & 26.9 & 22.8 & 25.2 & 0.4478$\pm$0.0177\\
    {Qwen2.5-VL-7B-Instruct}+Reflection~\cite{selfcorrection} & Reasoning  & 26.7 & 40.1 & 33.4 & 0.5826$\pm$0.0124\\
    {R1-OneVision-7B}~\cite{internvl} & Reasoning  & 22.9 & 30.7 & 30.2 & 0.5098$\pm$0.0099\\
    {Mulberry-Qwen2-VL-7B}~\cite{mulberry} & Reasoning  & 22.6 & 29.2  & 24.4 & 0.4740$\pm$0.0147\\
    {InternVL-2.5-8B}~\cite{internvl} & General  & 20.8 & 31.9  & 26.4 & 0.4838$\pm$0.0156\\
    {Llama-3.2-Vision-11B}~\cite{llama3}  & General  & 18.7 & 26.9  & 22.3 & 0.4265$\pm$0.0141\\
    {Llava-CoT-11B}~\cite{llavacot}  & Reasoning  & 17.4 &  26.9 & 22.4 & 0.4267$\pm$0.0140\\
    {{\method}-7B (Ours)} & Reasoning  & \bf 37.1 & \bf 43.3  & \bf 38.3 & \bf 0.6568$\pm$0.0179   \\
    \hline
   \multicolumn{6}{c}{\emph{Open-sourced $\sim$3B MLLMs}} \\
    \hline
    {Qwen2.5-VL-3B-Instruct}~\cite{qwen25vl} & General  & 18.8 & 23.1 & 18.8 & 0.3422$\pm$0.0244\\
    {Phi-3.5-Instruct}~\cite{phi3} & General  & 12.9 &  16.6 & 13.8 & 0.3181$\pm$0.0161\\
    {{\method}-3B (Ours)} & Reasoning  & \bf 29.4 & \bf 38.9  & \bf 34.5 & \bf 0.5840$\pm$0.0216   \\
   \bottomrule
\end{tabular}
}
}
\label{table:exp_comp}
\end{table}

\begin{figure}[h]
\centering
\begin{minipage}[t]{0.48\textwidth}
   \renewcommand\arraystretch{1.3}
   \centering
   \includegraphics[width=0.8\textwidth]{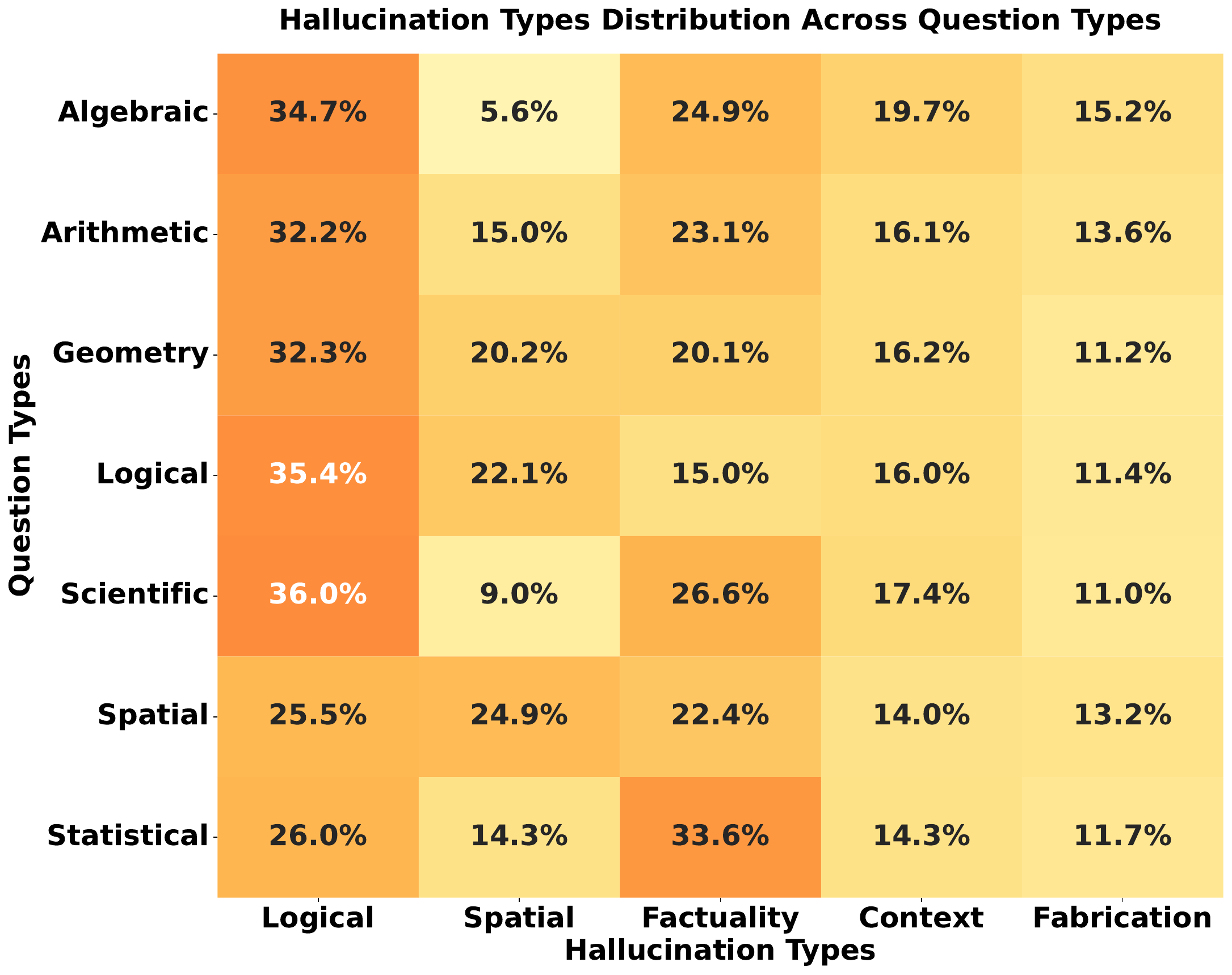}
   \caption{Distribution between question types and reasoning hallucination types.
   }
   \label{fig:question_hallu_correlation}
\end{minipage}\hspace{4mm}
\begin{minipage}[t]{0.48\textwidth}
   \renewcommand\arraystretch{1.3}
   \centering
   \includegraphics[width=0.8\textwidth]{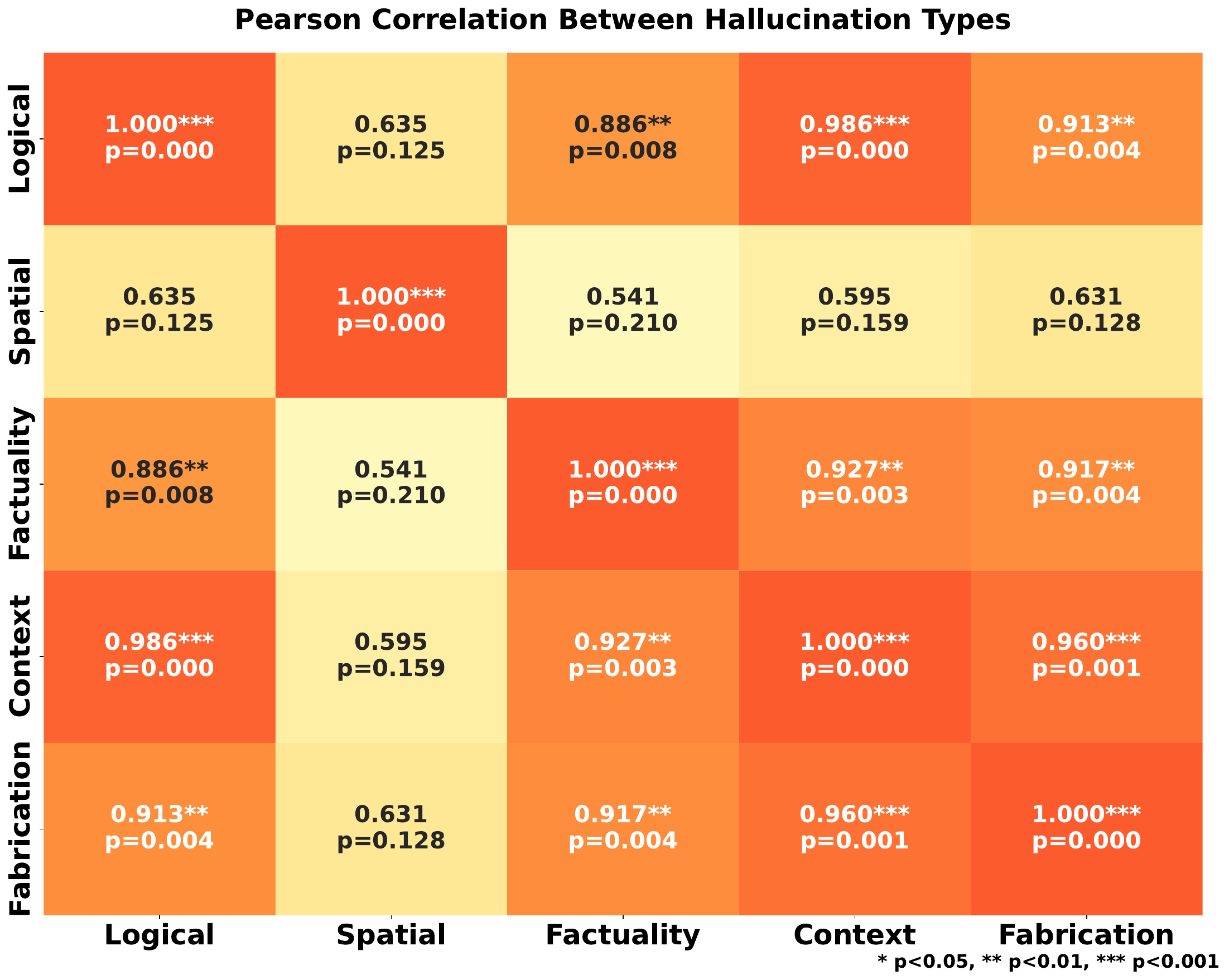}
   \caption{Pearson correlation among reasoning hallucination types.
   }
   \label{fig:hallu_correlation}
\end{minipage}
\end{figure}

\subsection{Online Reward Filtration}
While CRFT effectively controls data difficulty, it may still encounter training samples where all generated responses receive identical rewards, disrupting the optimization process. To address this without compromising training efficiency, we integrate offline data filtration~\cite{limo} into our approach, forming Online Reward Filtration (ORF).
In each iteration, for given question $\mathbf{x}$ with $G$ sampled responses ${\mathbf{r}_{1},...,\mathbf{r}_{G}}$, {\method} first computes the corresponding rewards ${r_{1},...,r_{G}}$ using the predefined reward function $\mathcal{R}$. If all responses share the same reward ($r_{1} = ... = r_{G}$), the question is discarded for that iteration, ensuring only diverse, meaningful samples contribute to optimization.
%

\subsection{Collaborative Hint Inference}
To further reduce reasoning hallucinations beyond training, we introduce Collaborative Hint Inference (CHI), which leverages an auxiliary LLM $\phi$ to provide context-specific guidance.
\begin{table}[h]
\caption{Probability of different types of reasoning hallucinations for each model.}\label{table:hallu_model_relation}
\centering
\setlength{\tabcolsep}{10pt} 
	\renewcommand{\arraystretch}{2.0}
	  		{ \fontsize{8.3}{3}\selectfont{
\begin{tabular}{l|ccccc}
\toprule
\bf Model & \bf Logical & \bf Spatial & \bf Factuality & \bf Context & \bf Fabrication \\
\midrule
Gemini-2-flash & 54.66\% & 29.33\% & 39.03\% & 28.48\% & 22.32\% \\
Qwen2.5-VL-7B & 68.94\% & 35.35\% & 51.07\% & 30.98\% & 23.91\% \\
Gemini-2-flash-thinking & 47.88\% & 25.44\% & 32.85\% & 22.51\% & 18.04\% \\
Virgo-72B & 63.97\% & 29.18\% & 40.65\% & 32.26\% & 21.32\% \\
QvQ-72B-Preview & 73.37\% & 37.93\% & 47.91\% & 47.52\% & 30.19\% \\
\bottomrule
\end{tabular}
}}
\end{table}
\begin{table}[h]
\centering
\begin{minipage}[t]{0.48\textwidth}
   \renewcommand\arraystretch{1.3}
   \centering
   \caption{Manually fixing reasoning chains experimental results on 10\% sampled questions with reasoning hallucination. 
   }
   \setlength{\tabcolsep}{3pt} 
	\renewcommand{\arraystretch}{1.7}
	  		{ \fontsize{8.3}{3}\selectfont{
   \begin{tabular}{lc|c}
      \toprule
      \bf Model & \bf Fix Reasoning & \bf Accuracy  \\
      \midrule
      GPT-4o & -  & 12.1  \\
      GPT-4o & \Checkmark & 68.5 \\
      \midrule
      Qwen2.5-VL-72B & -  & 10.4  \\
      Qwen2.5-VL-72B & \Checkmark & 72.4 \\
      \bottomrule
   \end{tabular}}}
   \label{table:manual_fixed_exp}
\end{minipage}\hspace{4mm}
\begin{minipage}[t]{0.48\textwidth}
   \renewcommand\arraystretch{1.3}
   \centering
   \caption{Correlation matrix of all metrics. All correlations are significant at $p < 0.001$ ($^{***}$).
   }
   \setlength{\tabcolsep}{3pt} 
	\renewcommand{\arraystretch}{3.1}
	  		{ \fontsize{8.3}{3}\selectfont{
   \begin{tabular}{l|llll}
\toprule
\bf{Metric}& \bf{Accuracy} & {$F_{\text{step}}$} & {$F_{\text{claim}}$} & \bf{LHS} \\
\midrule
\bf{Accuracy} & 1.000 & 0.864$^{***}$ & 0.918$^{***}$ & 0.889$^{***}$ \\
$F_{\text{step}}$ & 0.864$^{***}$ & 1.000 & 0.975$^{***}$ & 0.915$^{***}$ \\
$F_{\text{claim}}$ & 0.918$^{***}$ & 0.975$^{***}$ & 1.000 & 0.933$^{***}$ \\
\bf{LHS} & 0.889$^{***}$ & 0.915$^{***}$ & 0.933$^{***}$ & 1.000 \\
\bottomrule
\end{tabular}
   }}
   \label{table:ablation_metric_corre}
\end{minipage}
\end{table}
Given a question $\mathbf{x}$, CHI first uses a predefined question classification prompt to guide $\phi$ in predicting the question type $\mathbf{c}$. Based on this type, CHI generates two structured hints: a topic-specific hint $\mathbf{h}{\text{topic}}$, reflecting the general approach for the given type $\mathbf{c}$, and a question-specific hint $\mathbf{h}{\text{question}}$, tailored to the particular content of $\mathbf{x}$.
During inference, we generate response by $\mathbf{r} = \pi([\mathbf{h}_{\text{topic}}, \mathbf{x}, \mathbf{h}_{\text{question}}])$. The optimized MLLM can benefit from CHI and generate more accurate chains than vanilla MLLMs.

%% file: sections/experiments.tex
\section{Experiments}\label{sec:exps}

We state the full experimental setup of {\dataset} evaluation and training of {\method} in App.~\ref{sec:exp_implement}. And in the following, we illustrate the insightful findings on {\dataset} and the effectiveness of {\method}.
\subsection{Empirical Analysis}\label{sec:exp_analysis}
\noindent\textbf{Overall results. }
As shown in Table~\ref{table:exp_comp}, O1 achieves the highest accuracy at 49.7, outperforming Gemini-2-flash-thinking by 2.1. However, O1 scores lower in factuality, likely due to generating shorter, more summarized reasoning chains that reduce step coverage and recall.
For most models except GPT-4o and O1, step scores exceed claim scores, suggesting that generating coarse reasoning steps is easier than detailed calculation steps. The LHS scores align with step scores, confirming the reliability of {\dataset}'s metrics.
Focusing on open-source Qwen-VL models, increasing parameters from 3B to 72B raises accuracy from 18.8 to 38.8. 
Moreover, better pretraining in Qwen2.5-VL improves both accuracy and factuality/LHS, indicating that enhanced pretraining reduces hallucinations.
%

\noindent\textbf{Consistency among evaluation metrics. }
As shown in Table~\ref{table:ablation_metric_corre}, we compute pearson correlation coefficients between \textit{Accuracy}, $F_{\text{step}}$, $F_{\text{claim}}$, and LHS across all models. All pairs of metrics exhibit very strong positive correlations ($r=0.86$--$0.98$), with all correlations being highly significant ($p<0.001$). These results indicate that the hallucination rate in the reasoning chains has large correlation with final answers, and inspire our hallucination mitigation method. 

\noindent\textbf{Correlation between reasoning hallucination and final accuracy. } 
%
Given correlations between reasoning chains and answer accuracy, we conduct preliminary study to show the impact of hallucination mitigation. We manually corrected the reasoning chains for about 10\% of commonly misanswered questions by GPT-4o and Qwen2.5-VL-72B, then prompt the models to reconsider their final answers. As shown in Table~\ref{table:manual_fixed_exp}, this correction significantly improves the answer accuracy to around 70\%, confirming that reducing reasoning hallucinations directly enhances overall model performance. 

\begin{table}[h]
\caption{Hallucination type rates in MIRAGE benchmark questions of Qwen-7/72B with different pertaining data. Pretraining with higher quality data leads to less logical, fabrication, and factual hallucinations.}\label{table:hallu_pretrain_data_relation}
\centering
\setlength{\tabcolsep}{10pt} 
	\renewcommand{\arraystretch}{4.0}
	  		{ \fontsize{8.3}{3}\selectfont{
\begin{tabular}{l|ccccc}
\toprule
\bf Model & \bf Logical & \bf Factuality & \bf Spatial & \bf Context & \bf Fabrication \\
\hline
Qwen2.5-VL-72B & 47.7\% & 33.7\% & 29.2\% & 21.6\% & 16.5\% \\
Qwen2-VL-72B & 59.3\% & 45.4\% & 32.7\% & 32.6\% & 26.5\% \\
\hline
Qwen2.5-VL-7B & 64.7\% & 45.7\% & 33.4\% & 29.3\% & 25.5\% \\
Qwen2-VL-7B & 74.0\% & 60.6\% & 35.6\% & 42.7\% & 35.4\% \\
\bottomrule
\end{tabular}
}}
\end{table}

\begin{table}[h]
\caption{Hallucination type rates in MIRAGE benchmark questions of Qwen-2.5-VL. Larger Models lead to less logical, fabrication, and factual hallucinations.}\label{table:hallu_model_size_relation}
\centering
\setlength{\tabcolsep}{10pt} 
	\renewcommand{\arraystretch}{4.0}
	  		{ \fontsize{8.3}{3}\selectfont{
\begin{tabular}{l|ccccc}
\toprule
\bf Model & \bf Logical & \bf Factuality & \bf Spatial & \bf Context & \bf Fabrication \\
\hline
Qwen2.5-VL-72B & 47.7\% & 33.7\% & 29.2\% & 21.6\% & 16.5\% \\
Qwen2.5-VL-7B & 64.7\% & 45.7\% & 33.4\% & 29.3\% & 25.5\% \\
Qwen2.5-VL-3B & 78.9\% & 60.1\% & 36.7\% & 37.9\% & 38.1\% \\
\bottomrule
\end{tabular}
}}
\end{table}

\noindent\textbf{Relation between pretraining data and hallucination types. }
We also explore relations between pretraining data and hallucination types. Specifically, we keep use Qwen-VL~\cite{qwen2vl,qwen25vl} with different pretraining data (\emph{i.e.}, Qwen2-VL and Qwen2.5-VL) and compare the hallucination rates of each hallucination type. As shown in Table~\ref{table:hallu_pretrain_data_relation}, Qwen2.5-VL models have less logical, factual, and fabrication hallucination rates than those of Qwen2-VL models. A possible explanation is that pretraining data with higher quality provides more accurate factual knowledge and reasoning chains to models, such that models can avoid logical and factuality hallucinations during inference. 
Nevertheless, the spatial hallucination does not significantly reduced, which indicates that current MLLMs still show weak visual reasoning capabilities. 

\noindent\textbf{Relation between model size and hallucination types. }
We also explore relations between pretraining data and hallucination types. Specifically, we keep use Qwen2.5-VL~\cite{qwen25vl} with different model sizes (\emph{i.e.}, 3B/7B/72B) and compare the hallucination rates of each hallucination type. As shown in Table~\ref{table:hallu_model_size_relation}, Larger Qwen2.5-VL models have less logical, factual, and fabrication hallucination rates than those of smaller models. A possible explanation is that models owning more model parameters have more capabilities for accurate factual knowledge and reasoning chains to models, such that models can avoid logical and factuality hallucinations during inference. 
Nevertheless, the spatial hallucination does not significantly reduced, which indicates that current MLLMs still show weak visual reasoning capabilities. 

\noindent\textbf{Correlation between question and hallucination types. }
We also analyze the relationship between question types and hallucination patterns, as shown in Fig.~\ref{fig:question_hallu_correlation}. Results indicate that logical hallucinations are widespread across various question types, while certain hallucination types are more closely associated with specific question types.
Specifically, logical and spatial hallucinations are particularly common in logical questions, reflecting the high demands for complex reasoning and visual transformations that current MLLMs struggle with~\cite{mmiq}. In contrast, statistical and scientific questions tend to exhibit more factuality hallucinations, likely due to their reliance on precise knowledge retrieval.
These findings highlight the specific vulnerabilities of MLLMs in handling diverse reasoning tasks.

\begin{table}[h]
\centering
   \renewcommand\arraystretch{1.3}
   \centering
   \caption{Ablation study of {\method}-7B, where CHI means collaborative hint inference. 
   }
   \setlength{\tabcolsep}{2pt} 
	\renewcommand{\arraystretch}{2.0}
	  		{ \fontsize{8.3}{3}\selectfont{
   \begin{tabular}{l|cccc|cccc|c}
      \toprule
       \bf Method & \bf GRPO & \bf CRFT & $\mathbf{h}_{\text{topic}}$ & $\mathbf{h}_{\text{question}}$ &  \bf Accuracy & $F_{\text{step}}$ & $F_{\text{claim}}$ & \bf LHS & \bf MathVista \\
      \midrule
      Qwen2.5-VL-7B & \XSolidBrush & \XSolidBrush & \XSolidBrush & \XSolidBrush & 28.8 & 34.7 & 31.7 & 0.5996 & 68.2   \\
      +GRPO & \Checkmark & \XSolidBrush & \XSolidBrush & \XSolidBrush & 33.7 & 41.0 & 35.9 & 0.6180 & 70.7 \\
      +CRFT & \Checkmark & \Checkmark & \XSolidBrush & \XSolidBrush & 35.7 & 41.8 & 37.3 & 0.6193 & 71.9 \\
      
      +$\mathbf{h}_{\text{topic}}$ & \Checkmark & \Checkmark & \Checkmark & \XSolidBrush & 36.2 & 42.6 & 37.6 & 0.6335 & 72.2 \\
      +$\mathbf{h}_{\text{question}}$ & \Checkmark & \Checkmark & \XSolidBrush & \Checkmark & 36.5 & 42.2 & 37.6 & 0.6224 & 72.2 \\
      +full CHI & \XSolidBrush & \XSolidBrush & \Checkmark & \Checkmark & \bf 37.1 & \bf 43.3 & \bf 38.3 & \bf 0.6568 & \bf 72.3 \\
      Qwen2.5-VL-7B + full CHI & \XSolidBrush & \XSolidBrush & \Checkmark & \Checkmark & 29.0 & 34.9 & 32.1 & 0.6011 & 68.3 \\
      \bottomrule
   \end{tabular}}}
   \label{table:ablation_component}
\end{table}

\noindent\textbf{Correlation among hallucination types. }
We further analyze correlations among hallucination types using pearson coefficients. As shown in Fig.~\ref{fig:hallu_correlation}, logical hallucinations strongly correlate with factuality, context, and fabrication errors, likely because flawed logic often leads to context inconsistency and factual errors.
Notably, spatial hallucinations, which arise from complex visual operations, show relatively low correlation with other hallucinations, suggesting they are more independent and unique to multimodal models rather than text-based LLMs.
These findings highlight the need for targeted mitigation strategies for hallucination types, particularly for challenging spatial reasoning errors.


\noindent\textbf{Hallucination rate comparison across models. }
To quantitatively assess the impact of model design and training on reasoning hallucinations, we analyzed five representative MLLMs, as shown in Table~\ref{table:hallu_model_relation}.
QvQ-72B-Preview exhibits the highest overall hallucination rates, especially in Logical (73.37\%) and Context (47.52\%) categories, significantly higher than Virgo-72B, which shares the same base model but benefits from more effective fine-tuning.
In contrast, Gemini-2-flash-thinking consistently shows the lowest hallucination rates, particularly in Logical (47.88\%), Spatial (25.44\%), and Fabrication (18.04\%) categories, indicating superior robustness.
%

\noindent\textbf{Existing solutions are not sufficient to mitigate hallucination. }Training-free methods like self-reflection~\cite{selfcorrection} and visual inference chain~\cite{vic} generally degrade both accuracy and LHS on base models without sufficient reasoning capabilities (Table~\ref{table:exp_comp}), highlighting their limitations. Similarly, SFT-based methods can improve hallucination mitigation on larger models (e.g., Virgo-72B) but often fail to enhance smaller models, suggesting that model capacity plays a critical role in the effectiveness of external supervision. More detailed analysis can be found in App.~\ref{sec:app_more_analysis}.


\subsection{Empirical Analysis of {\method}}\label{sec:exp_ablation}
We use {\method}-7B as an example and conduct an ablation study on both {\dataset} and a standard benchmark, MathVista~\cite{mathvista}.
More in-depth analysis can be found in App.~\ref{sec:app_more_analysis}.

\begin{table}[h]
\caption{Hallucination type rates in MIRAGE benchmark questions of Qwen2.5-VL-3B/7B and corresponding Logos-3B/7B. Our proposed method leads to less logical and fabrication hallucinations.}\label{table:hallu_logos_relation}
\centering
\setlength{\tabcolsep}{10pt} 
	\renewcommand{\arraystretch}{4.0}
	  		{ \fontsize{8.3}{3}\selectfont{
\begin{tabular}{l|ccccc}
\toprule
\bf Model & \bf Logical & \bf Factual & \bf Spatial & \bf Context & \bf Fabrication \\
\hline
Qwen2.5-VL-7B & 64.7\% & 45.7\% & 33.4\% & 29.3\% & 25.5\% \\
Logos-7B & 49.3\% & 39.7\% & 29.9\% & 23.8\% & 15.6\% \\
\hline
Qwen2.5-VL-3B & 78.9\% & 60.1\% & 36.7\% & 37.9\% & 38.1\% \\
Logos-3B & 57.1\% & 47.4\% & 36.7\% & 31.8\% & 24.0\% \\
\bottomrule
\end{tabular}
}}
\end{table}

\noindent\textbf{Comparison with previous methods. }
We compare {\method}-7B with other 7B models to validate its effectiveness.
Compared to the base model Qwen2.5-VL-7B\cite{qwen25vl}, {\method}-7B achieves an 8.3 gain in accuracy, and outperforms the base by 8.6 on $F_{\text{step}}$ and 6.6 on $F_{\text{claim}}$, approaching the performance of the larger Virgo-72B\cite{virgo}.
These results, consistent with LHS scores, indicate that {\method} effectively reduces reasoning hallucinations, improving reliability across reasoning chains.
Similar gains are also observed for {\method}-3B, showing the compatibility of our framework across different model scales.

\noindent\textbf{Whether Logos reduces reasoning hallucination or not. }
Finally we investigate the hallucination mitigate effect on each hallucination type. As shown in Table~\ref{table:hallu_logos_relation}, Logos-7B reduces logical hallucination by 15.4\% and fabrication hallucination by 10\%. Similar results can also be found in Logos-3B. Nevertheless, we do not find significant halucination mitigation on spatial and factuality hallucination on both Logos models. A possible reason is that reinforcement learning does not introduce new knowledge and only refines the logic of reasoning chains. 

\noindent\textbf{Ablation Study of Each Component of {\method}. }
We first investigate the effect of each key component in {\method}. The experimental results are shown in Table~\ref{table:ablation_component}. After adopting reinforcement learning on the base model, the accuracy on {\dataset} and MathVista achieves 33.7 and 70.7 respectively. benefiting from RL, the step score and claim score also increase to 41.0 and 35.9. 
After adopting CRFT, both accuracy and $F_{\text{claim}}$ further increases to 35.7 and 37.3 respectively. 
By further integrating CHI, {\method}-7B achieves 37.1 on {\dataset} and 72.3 on Mathvista. 
Note that directly adopt CHI on base model does not lead to performance improvement, which further proves the findings in Sec.~\ref{sec:exp_analysis}. 

\noindent\textbf{How {\method} mitigates reasoning hallucination? } 
We evaluated the impact of CRFT by comparing the accuracy of {\method}-7B on 8-round sampling across training dataset, before and after training. The accuracy increases from 24.8\% to 68.3\%, indicating that CRFT effectively guides the model to generate correct reasoning chains.
Meanwhile, we calculate the ``Logical'' hallucination rate in Qwen2.5-VL-7B and Logos-7B, which reduces from 57.1 to 49.3. This result shows that CRFT benefits to encourage model learning logic-consistent reasoning chains to mitigate hallucination.

%% file: sections/conclusion.tex
\vspace{-0.3em}
\section{Conclusion}\label{sec:conclusion}
\vspace{-0.2em}
We propose {\dataset}, which isolates reasoning hallucinations by questions where inputs are correctly perceived but reasoning errors persist. 
For analysis of reasoning hallucination, {\dataset} proposes multi-level evaluation metrics, covering different levels of the reasoning chains. 
%
Our findings reveal that the model scale, data scale, and training stages of MLLMs: (1) significantly influence the degree of logical, fabrication, and factual hallucinations; (2) show no effective improvement on spatial hallucinations caused by misinterpretations of spatial relationships, suggesting that current MLLMs exhibit weak visual reasoning capabilities and struggle to benefit from simple scaling of training resources; and 3) correlations between question types and specific reasoning hallucination patterns, highlighting critical challenges and mitigation for specific types. These findings will provide insights for future MLLM development. 
To address this, we propose {\method}, a method using curriculum reinforcement fine-tuning and collaborative hint inference to reduce logical hallucination for higher accuracy. {\method} provides a baseline and offers insights for reducing hallucinations.

%% file: sections/appendix.tex
\clearpage
\appendix

This appendix mainly contains:
\begin{itemize}
\item Hallucination type definition in Section \ref{sec:app_hallu_def}
\item Detailed experimental settings in Section \ref{sec:exp_implement}
\item Additional quantitative results in Section \ref{sec:quantitative}
\item Additional in-depth analysis in Section \ref{sec:app_more_analysis}
\item Dataset examples in Section \ref{sec:dataset_example}
\item More qualitative results in Section \ref{sec:qualitative}
\item Statement of limitations in Section \ref{sec:limitation}
\item Statement of broader impact in Section \ref{sec:impact}
\end{itemize}

\section{Hallucination Type Definition}\label{sec:app_hallu_def}

In this section, we summarize the hallucination types mentioned in Sec.~\ref{sec:evaluation_factuality}, which are listed in Table~\ref{table:mm_hallu_type}.
Specifically, the multimodal reasoning hallucinations can be categorized into five distinct types, \emph{i.e.}, spatial hallucination, logical hallucination, factuality hallucination, context hallucination, and fabrication hallucination. The detailed descriptions are summarized in Table~\ref{table:mm_hallu_type}.
\begin{table}
    \centering
    \caption{Categories of multimodal reasoning hallucination investigated in {\dataset}. }
\setlength{\tabcolsep}{1.5pt} %
    \renewcommand{\arraystretch}{4.0}%
    { \fontsize{8.3}{3}\selectfont{
\begin{tabular}{l|c}
   \Xhline{1.5pt}
   {\bf{Hallucination Type}} & {\bf{Hallucination Description}}\\
   \Xhline{1.5pt}
   
   Spatial Hallucination & Errors in reasoning about spatial relationships, shapes, or complex visual operations.  \\
   Logical Hallucination & Errors in logical consistency or reasoning, even when surface-level facts are correct. \\
   Factuality Hallucination & Factually incorrect claims about scientific principles or established knowledge in input data. \\
   Context Hallucination & Inconsistencies between intermediate reasoning steps and final predictions. \\
   Fabrication Hallucination & Entirely invented values, entities, or relationships not in input data or real world. \\
   \Xhline{1.5pt}
\end{tabular}
}
}
\label{table:mm_hallu_type}
\end{table}
\section{Experimental Details of {\dataset} and {\method}}\label{sec:exp_implement}
\subsection{Experimental Setup}
\noindent\textbf{Implementation Details. }During {\dataset} evaluation, we leverage GPT-4o~\cite{gpt-4o} to judge the accuracy of final answers. And for factuality and LLMs hallucination score metrics, to reduce the cost while keeping comparable evaluation accuracy, we utilize DeepSeek-V3~\cite{deepseek-v3} for both metrics, and utilize Qwen2.5-72B-Instruct~\cite{qwen2.5} as well as Llama-3.1-70B-Instruct~\cite{llama3} for LLMs hallucination score. 
During training of {\method}, we use Qwen2.5-VL-7B-Instruct~\cite{qwen25vl} as the base model. The visual encoder is frozen to avoid catastrophic forgetting of visual perception ability~\cite{huo2025continue,dengzero,dong2024mr}. During training, we collect 13K mathematical questions with K12-level difficulty and $\sim$1K text-only math questions from LIMO~\cite{limo} as training data.
The batch size is 128. For each training sample, the rollout samples $G$ is 8 by default. The initial learning rate is  $1\times 10^{-6}$, both warmup strategy and cosine learning rate scheduler are adopted to stabilize training. We optimize {\method} by 10 epochs using AdamW~\cite{adamw} during each stage. The number of CRFT stages is set to 1, and we will discuss this choice in Sec.~\ref {sec:exp_ablation}. Benefiting from the filtration mechanism in CRFT and ORF, the total training time is less than 24 hours.  
%
All programs are constructed by PyTorch~\cite{pytorch} toolkit and vLLM~\cite{vllm} framework. All the experiments are conducted on 8 NVIDIA RTX A6000 GPUs.

\subsection{Baseline Models }\label{sec:exp_baseline}
We evaluate various reasoning and CoT-enhanced general MLLMs, including black-box MLLMs~\cite{o1,gemini,gpt-4o} and open-sourced MLLMs~\cite{qwen25vl,qwen2vl,internvl,llama3,phi3}. We also analyze reasoning-enhanced methods~\cite{mulberry,llavacot,qvq,virgo,vic,selfcorrection} to explore the hallucination mitigation effectiveness.

\paragraph{Black-box MLLMs }include GPT-4o~\cite{gpt-4o} and O1~\cite{o1} from OpenAI, as well as Gemini-2-flash and Gemini-2-flash-thinking~\cite{gemini} from Google. 
These models have shown state-of-the-art reasoning or chain-of-thought thinking capabilities in various tasks. 

\paragraph{Open-sourced MLLMs }cover both specifically-designed reasoning MLLMs (\emph{e.g.}, QvQ-72B~\cite{qvq} and Virgo-72B~\cite{virgo}), and General MLLMs including Qwen2-VL~\cite{qwen2vl}, Qwen2.5-VL~\cite{qwen25vl}, InternVL-2.5~\cite{internvl}, Llama-3.2-Vision~\cite{llama3} and Phi-3.5-Instruct~\cite{phi3}. All these models have shown competitive inherent or chain-of-thought reasoning capabilities. 
Note that the parameter numbers of selected models are largely varied from 3B to 72B, ensuring that models with different scales can be analyzed in our experiments. 

\paragraph{Reasoning-enhanced General MLLMs. }Besides, to comprehensively evaluate the capabilities of reducing reasoning hallucination, we also assess multiple training-free and training-based hallucination mitigation methods, including self-reflection~\cite{selfcorrection}, question decomposition~\cite{vic}, and supervised fine-tuning~\cite{r1-onevision}. All these methods have shown effectiveness in improving reasoning capabilities. 

\subsection{Prompts Used in Construction and Evaluation}
\begin{figure}
\begin{center}
\includegraphics[width=\textwidth]{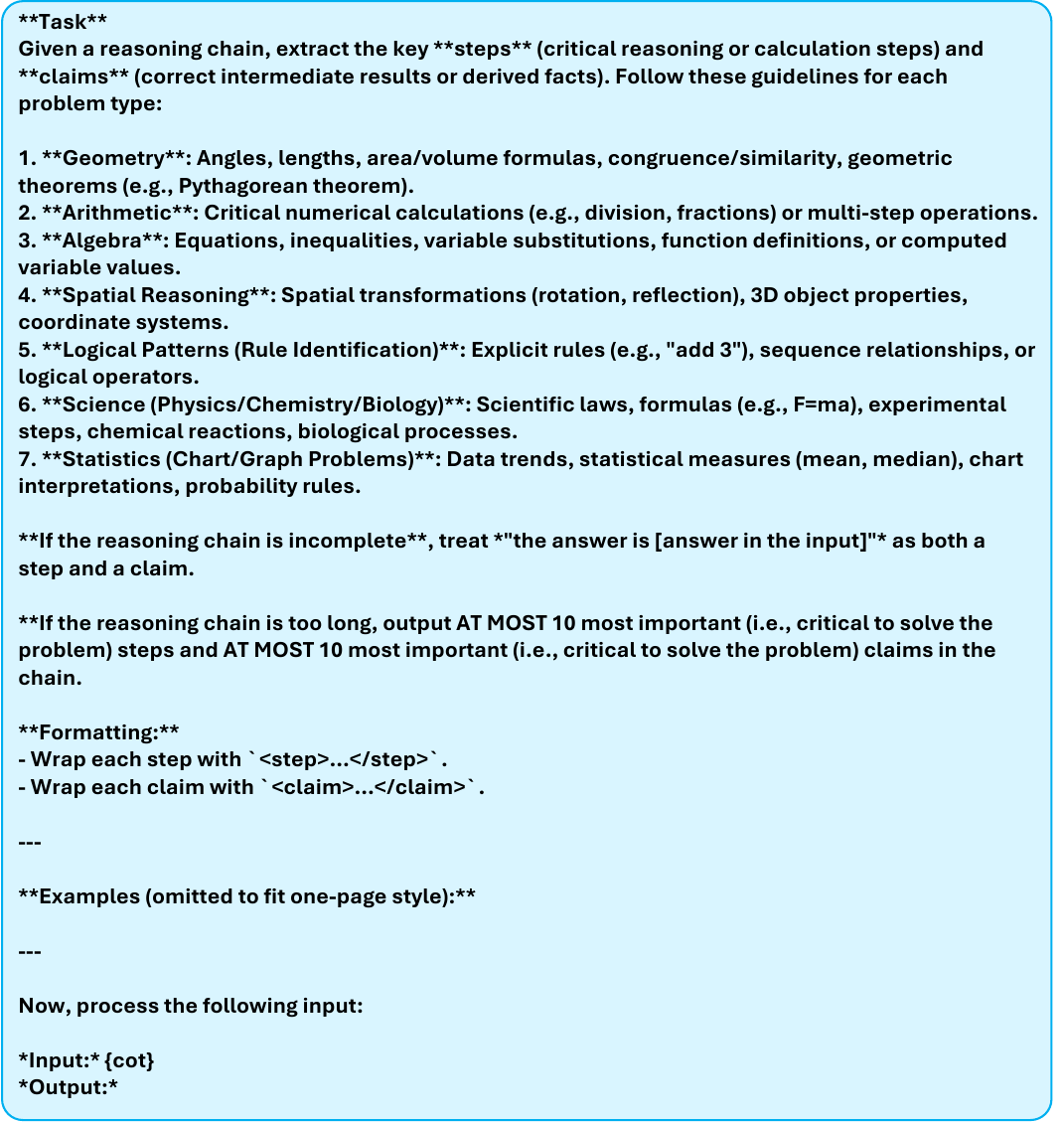}
\end{center}
\caption{The evaluation prompt used to extract intermediate results. 
}
\label{fig:extract_intermediate_prompt}
\end{figure}

\begin{figure}
\begin{center}
\includegraphics[width=\textwidth]{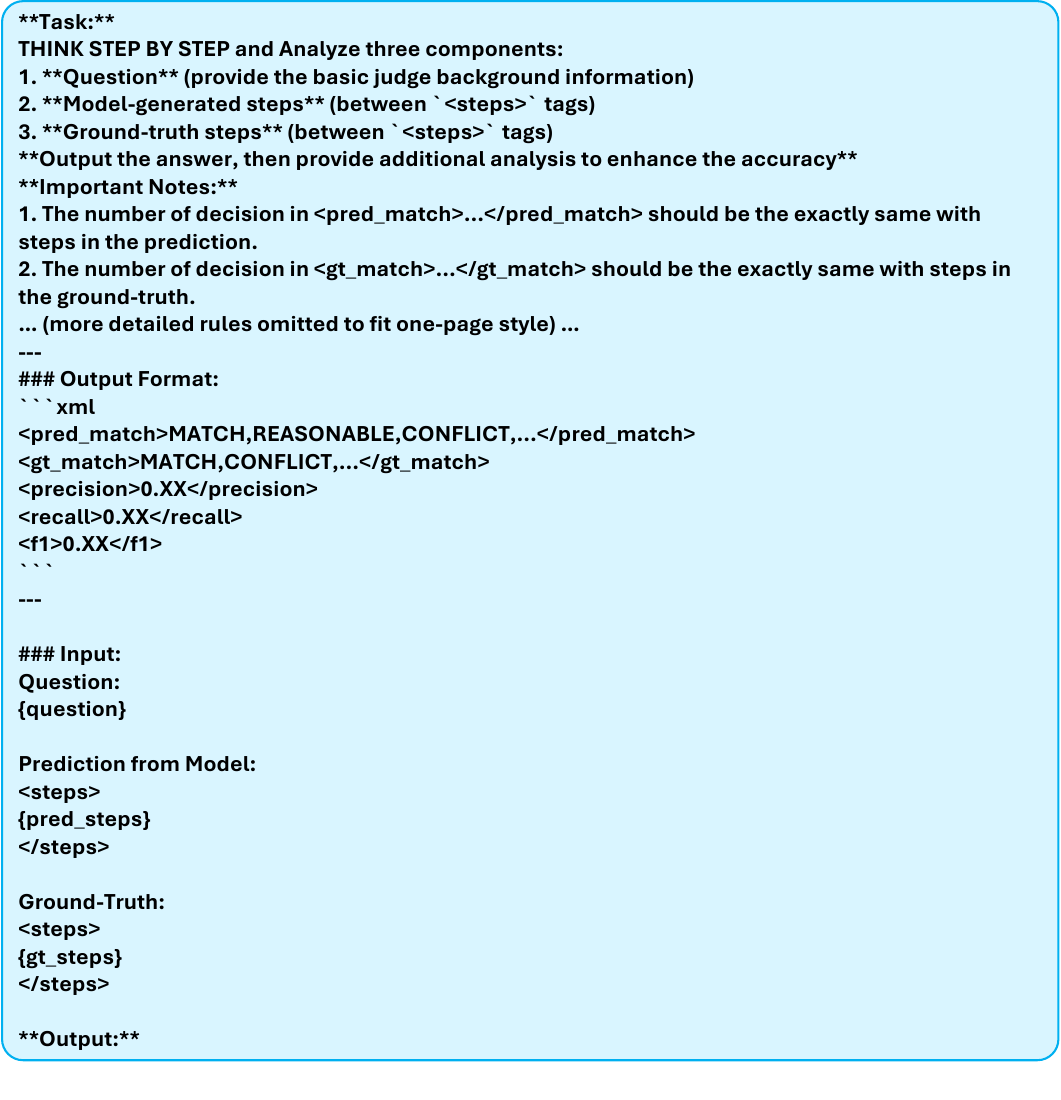}
\end{center}
\caption{The evaluation prompt used for factuality assessment (\emph{e.g.}, $F_{\text{step}}$). 
}
\label{fig:factuality_eval_prompt}
\end{figure}

\begin{figure}
\begin{center}
\includegraphics[width=\textwidth]{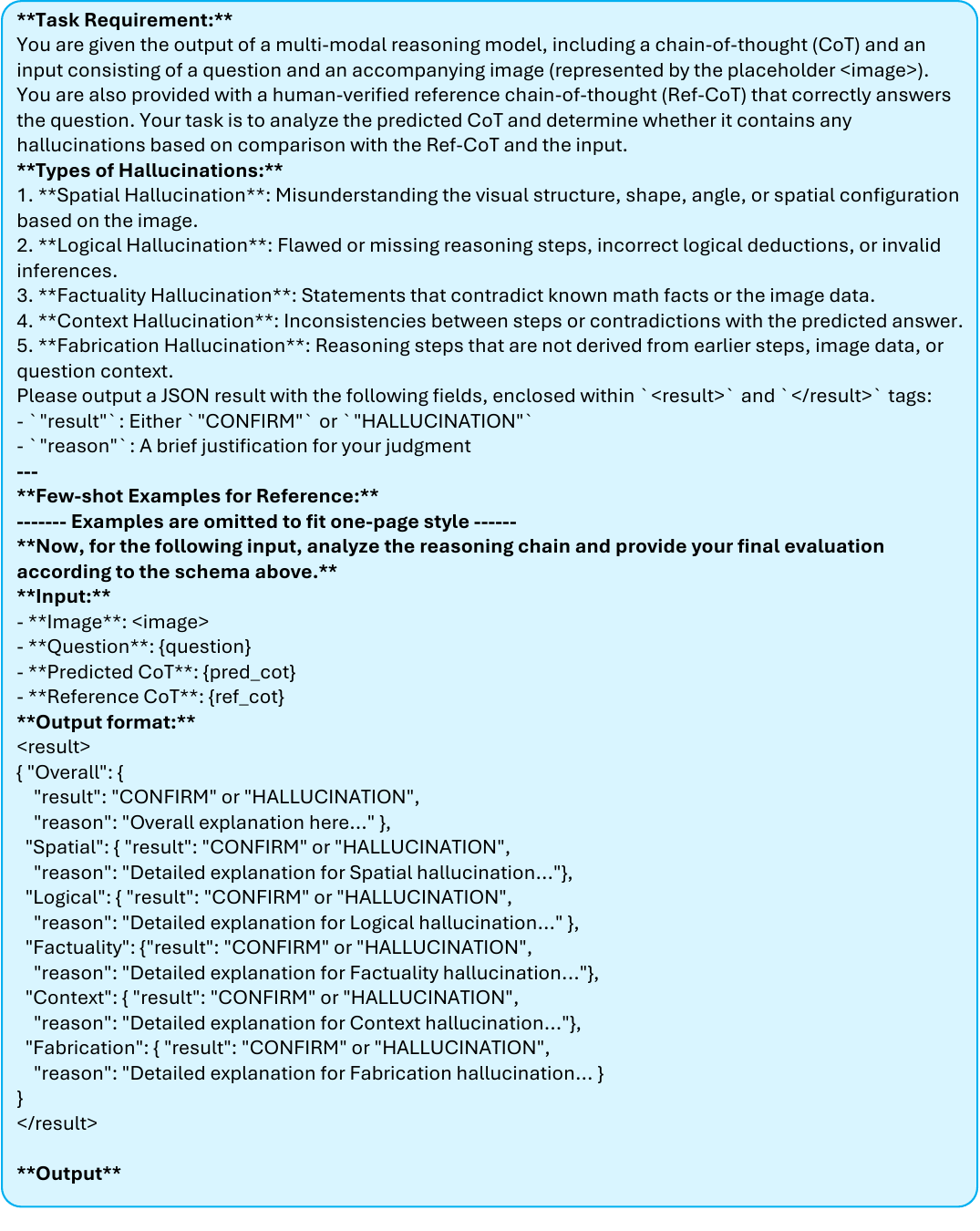}
\end{center}
\caption{The evaluation prompt used to detect hallucination types in reasoning chains. 
}
\label{fig:halludet_prompt}
\end{figure}

\begin{figure}
\begin{center}
\includegraphics[width=\textwidth]{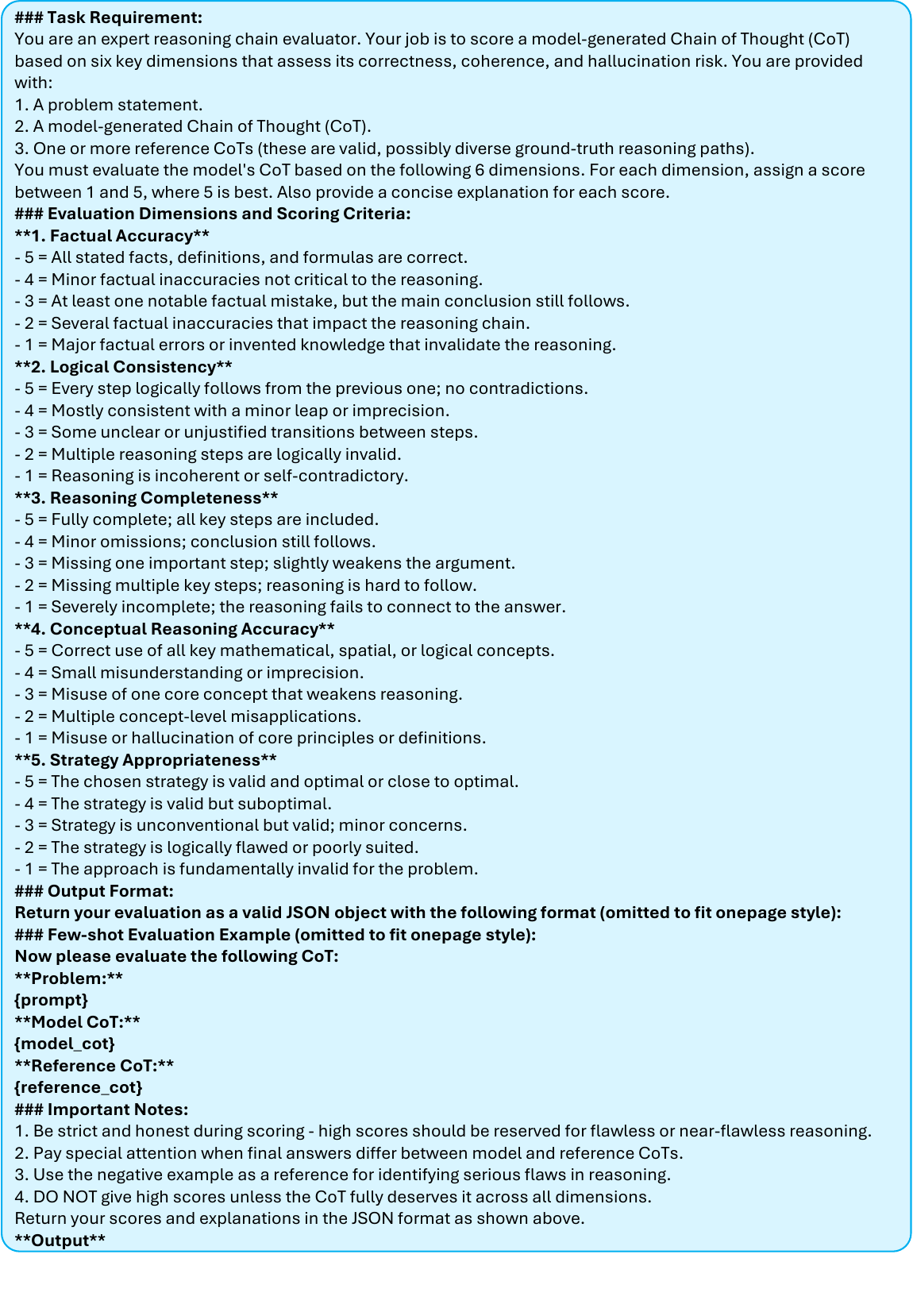}
\end{center}
\caption{The evaluation prompt used for LLMs hallucination score extraction. 
}
\label{fig:lhs_prompt}
\end{figure}

For the reproducibility of our work, we release the critical prompts used in {\dataset} construction and evaluation. 
Fig.~\ref{fig:extract_intermediate_prompt} shows the prompt for extracting intermediate results (\emph{i.e.}, steps and claims). Fig.~\ref{fig:factuality_eval_prompt} shows the prompt of intermediate results matching results (\emph{i.e.}, factuality evaluation prompt). Fig.~\ref{fig:halludet_prompt} shows the prompt used to detect specific reasoning hallucination types. And Fig.~\ref{fig:lhs_prompt} shows the prompt used to calculate the LHS score.
\subsection{Computational Resources and Time}
During the first stage of C-RFT, {\method}-7B uses 8 NVIDIA A6000 GPUs to train a 7B model, and the total training time is 16 hours. During the second stage (since the optimal stage number $k$ is 1, as discussed in App.~\ref{sec:app_more_analysis}), {\method}-7B uses the same 8 A6000 GPUs and requires 6 hours to complete training. During Inference, benefiting from the optimization of the vLLM~\cite{vllm} framework, {\method}-7B only requires one A6000 GPU for inference reasoning chains.
And both training and inference of {\method}-3B model require fewer computational resources. 
\subsection{Significance Computation}
To calculate the correlation and the corresponding significance value, we leverage scipy package and call \texttt{pearsonr} function to calculate pearson correlation coefficient with corresponding significance (\emph{i.e.}, p-value). 

\begin{table}[h]
   \renewcommand\arraystretch{1.3}
   \centering
   \caption{Effect of reinforcement learning algorithm. We remove CHI and directly assess the original output of each model.  
   }
    \vspace{0.5em}
   \setlength{\tabcolsep}{10.6pt} 
	\renewcommand{\arraystretch}{2.2}
	  		{ \fontsize{8.3}{3}\selectfont{
   \begin{tabular}{l|cc}
      \toprule
      RL & \textbf{MIRAGE}  & \textbf{MathVista} \\
      \midrule
      PPO & 29.9 & {69.3} \\
       DAPO & 30.0 & 69.6 \\
      GRPO & 35.7 & {71.9} \\
      \bottomrule
   \end{tabular}}}
   \label{table:ablation_rl_algo}
\end{table}

\begin{table}[h]
\centering
\begin{minipage}[t]{0.48\textwidth}
   \renewcommand\arraystretch{1.3}
   \centering
   \caption{Effect of online reward filtration.
   }
    \vspace{0.5em}
   \setlength{\tabcolsep}{7pt} 
	\renewcommand{\arraystretch}{2.2}
	  		{ \fontsize{8.3}{3}\selectfont{
   \begin{tabular}{l|cc}
      \toprule
      \bf ORF & \bf {\dataset} & \bf MathVista \\
      \midrule
      \XSolidBrush & 34.2 & 69.6  \\
      \Checkmark & 37.1 & 72.3 \\
      \bottomrule
   \end{tabular}}}
   \label{table:ablation_orf}
\end{minipage}\hspace{4mm}
\begin{minipage}[t]{0.48\textwidth}
   \renewcommand\arraystretch{1.3}
   \centering
   \caption{Effect of the KL-divergence in {\method}.
   }
   \setlength{\tabcolsep}{10.6pt} 
	\renewcommand{\arraystretch}{2.2}
	  		{ \fontsize{8.3}{3}\selectfont{
   \begin{tabular}{l|cc}
      \toprule
      \textbf{KL-Div} & \textbf{MIRAGE}  & \textbf{MathVista} \\
      \midrule
      1e-2 & 31.0 & 67.0 \\
      1e-3 & 35.0 & 70.0 \\
      1e-4 & 36.7 & 71.1 \\
      0 ({\method}) & 37.1 & 72.3 \\
      \bottomrule
   \end{tabular}}}
   \label{table:ablation_KL}
\end{minipage}
\end{table}

\begin{table}[h]
\centering
\begin{minipage}[t]{0.48\textwidth}
   \renewcommand\arraystretch{1.3}
   \centering
   \caption{Comparison between CRFT and vanilla RL with longer training epochs. 
   }
    \vspace{0.5em}
   \setlength{\tabcolsep}{3pt} 
	\renewcommand{\arraystretch}{2.2}
	  		{ \fontsize{8.3}{3}\selectfont{
   \begin{tabular}{lc|cc}
      \toprule
      \bf Method & \bf Total Epochs & \bf {\dataset} & \bf MathVista \\
      \midrule
      Vanilla RL & 20 & 35.5 & 71.4  \\
      CRFT & 10+10 & 37.1 & 72.3 \\
      \bottomrule
   \end{tabular}}}
   \label{table:curriculum_vs_longrl}
\end{minipage}\hspace{4mm}
\begin{minipage}[t]{0.48\textwidth}
   \renewcommand\arraystretch{1.3}
   \centering
   \caption{Effect of curriculum learning stage $k$. 
   }
   \setlength{\tabcolsep}{10.6pt} 
	\renewcommand{\arraystretch}{2.2}
	  		{ \fontsize{8.3}{3}\selectfont{
   \begin{tabular}{l|cc}
      \toprule
      $k$ & \textbf{MIRAGE}  & \textbf{MathVista} \\
      \midrule
      0 & 35.0 & {70.7} \\
      1 & 37.1 & {72.3} \\
      2 & 37.2 & {72.3} \\
      3 & 37.2 & {72.3} \\
      \bottomrule
   \end{tabular}}}
   \label{table:ablation_curriculum}
\end{minipage}
\end{table}

\section{Detailed Quantitative Results}\label{sec:quantitative}
In addition to reporting average accuracy and overall LHS score for each model, we also report accuracy in each question topic and LHS score in each dimension. The detailed per-topic accuracy comparison results are shown in Table~\ref{table:per_topic_accuracy}, and the detailed per-dimension accuracy comparison results are shown in Table~\ref{table:per_dim_lhs}. Generally, all models suffer from unpromising accuracy on logical and spatial questions, which indicate that existing models still do not obtain sufficient visual pattern and relation reasoning abilities. Meanwhile, mathematic reasoning and statistical question reasoning are usually perform well on state-of-the-art MLLMs.
As for the LHS score, the score of logical consistency and reasoning completeness of some unpromising models are relatively low, which indicates that previous models still struggle in insufficient reasoning capabilities and result in reasoning hallucination. 
These results reveal the the vulnerability in reasoning MLLMs.

\section{More Analysis}\label{sec:app_more_analysis}

\subsection{Qualitative Results of Manually Fixing Examples}
\begin{figure}
\begin{center}
\includegraphics[width=\textwidth]{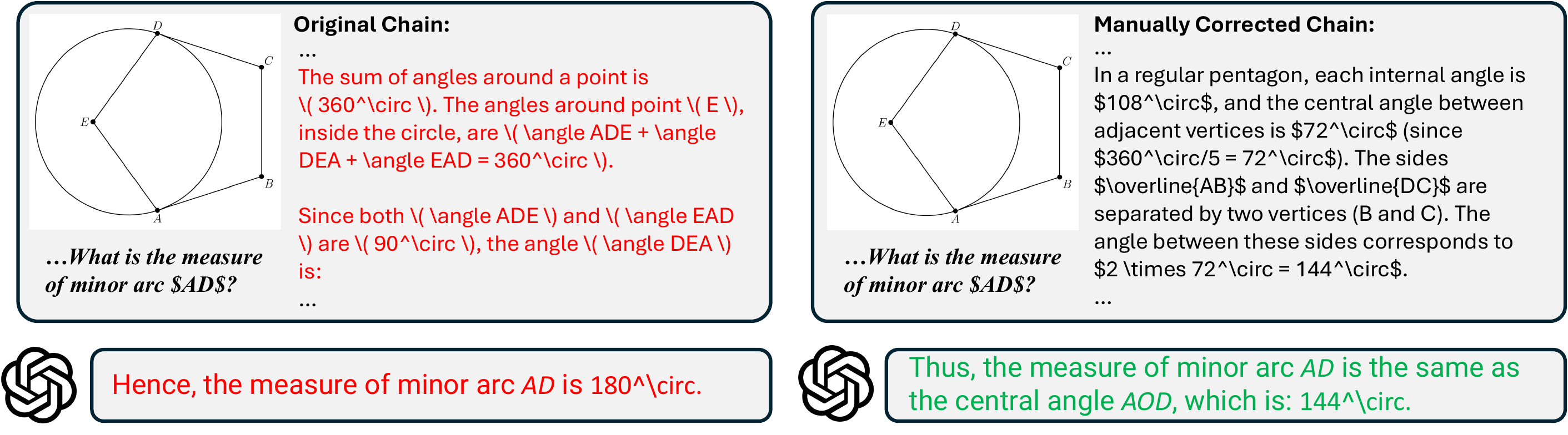}
\end{center}
\caption{Qualitative results of manually fixing the reasoning hallucination in the reasoning chain and inference the refined answers. The corrected reasoning chain (right) can guide MLLMs to predict correct answers, and the original reasoning chain with hallucination (left) still results in wrong answers. 
}
\label{fig:fixing_example}
\end{figure}

As stated in Sec.~\ref{sec:exp_ablation}, manually fixing hallucinations in the reasoning chains enhances overall model performance. In addition to the quantitative results, we also illustrate qualitative results from GPT-4o~\cite{gpt-4o}, which is shown in Fig.~\ref{fig:fixing_example}. The corrected reasoning chain (right) can guide MLLMs to predict correct answers, and the original reasoning chain with hallucination (left) still results in wrong answers.

\subsection{Pearson Correlation of Hallucination Type Among Single Model}
\begin{figure}
\begin{center}
\includegraphics[width=\textwidth]{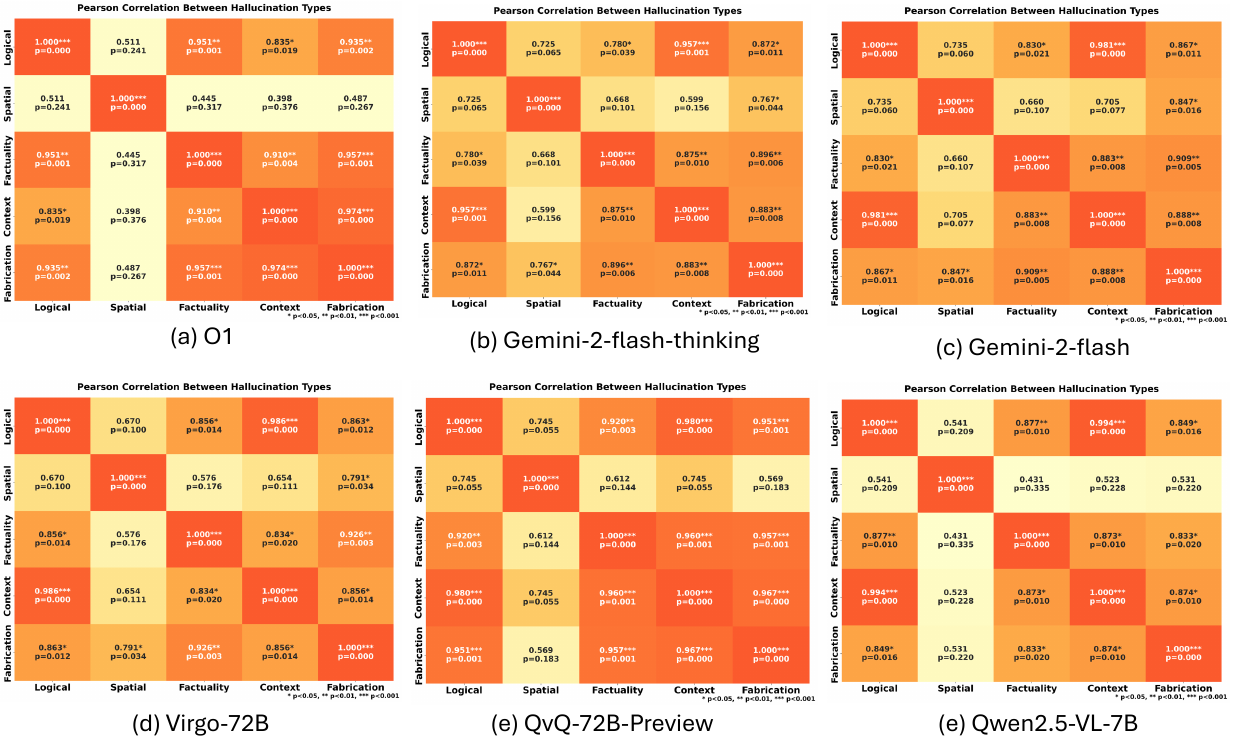}
\end{center}
\caption{Pearson correlation regarding hallucination types from six most representative MLLMs. All models tend to represent a similar pattern. 
}
\label{fig:pearson_single_model}
\end{figure}

In addition to stating the overall pearson correlation coefficient in Fig.~\ref{fig:hallu_correlation}, we also illustrate corresponding correlation from six most representative models, \emph{i.e.}, O1~\cite{o1}, Gemini-2-flash-thinking~\cite{gemini}, Gemini-2-flash~\cite{gemini}, Virgo-72B~\cite{virgo}, QvQ-72B-Preview~\cite{qvq}, and Qwen2.5-VL-7B~\cite{qwen25vl}. As shown in Fig~\ref{fig:pearson_single_model}, all models show a similar correlation pattern, which is consistent with Fig.~\ref{fig:hallu_correlation}. These results indicate the shared vulnerability in reasoning MLLMs. 

\subsection{Training-free Methods in {\dataset}.} We also explore some training-free methods to verify the hallucination mitigation capabilities. Therefore, we evaluate prompt-based self-reflection~\cite{selfcorrection} and visual inference chain~\cite{vic}. As shown in Table~\ref{table:exp_comp}, compared to Qwen2.5-VL-7B, both methods suffer performance degradation on both accuracy and LHS. These results indicate that for models without insufficient reasoning capabilities and hallucination-defending abilities, introducing training-free methods does not help to mitigate reasoning hallucination. This observation inspires us to integrate CHI into CRFT-enhanced MLLMs rather than the base model. 

\subsection{Supervised fine-tuning (SFT) Methods in {\dataset}} 
Intuitively, reasoning hallucination could be solved by further supervised fine-tuning to integrate the correct 
Hence, we select several SFT-based methods on 7B-level~\cite{mulberry,llavacot,r1-onevision} and 72B-level~\cite{qvq,virgo}, and evaluate on {\dataset}. As shown in Table~\ref{table:exp_comp}, on 72B MLLMs, introducing SFT can lead to better accuracy and partly mitigate reasoning hallucinations. Nevertheless, on 7B MLLMs, only Mulberry surpasses the base model by 3.1 on accuracy, while other methods do not lead to performance improvement and hallucination mitigation. This contradiction may come from the model's capacity for model sizes. Larger models with more inherent knowledge may be easier to mitigate hallucination via external supervision, while smaller models usually struggle with SFT.

\begin{table}[ht]
    \centering
    \caption{Accuracy comparison of each question topic in {\dataset}. }
\setlength{\tabcolsep}{0.5pt} %
    \renewcommand{\arraystretch}{3.5}%
    { \fontsize{8.3}{3}\selectfont{
\begin{tabular}{l|ccccccc|c}
   \Xhline{1.5pt}
   {\bf Model}  & {\bf Algebra} & {\bf{Arithmetic}} & {\bf{Geometry}} &  {\bf{Logical}} 
    &{\bf Scientific}&{\bf Spatial}&{\bf Statistical} & {\bf Overall}   \\
   \Xhline{1.5pt}
   
   
   \multicolumn{9}{c}{\bf{Black-Box MLLMs}} \\
    \hline
   Gemini-2-Flash-Thinking~\cite{gemini}  & 56.1 & 66.3 & 53.7 & 33.3 & 41.4 & 26.1 & 55.0  & 47.6  \\
   O1~\cite{o1}  & 50.9 & 64.1 & 60.0 & 37.7 & 42.5 & 37.8 & 51.9 & 49.7   \\
   Gemini-2-Flash~\cite{gemini}  & 51.2 & 57.6 & 50.2 & 34.1 & 32.3 & 26.7 & 55.8 & 44.1   \\
   GPT-4o~\cite{gpt-4o}  & 38.8 & 40.2 & 29.5 & 24.6 & 28.0 & 42.2 & 47.2  & 35.0 \\
   \hline
   \multicolumn{9}{c}{\bf{Open-sourced $\sim$72B MLLMs}} \\
    \hline
    {Qwen2.5-VL-72B-Instruct}~\cite{qwen25vl}  & 44.9 & 50.0 & 37.8 & 29.0 & 24.7 & 32.2 & 49.6  & 38.8 \\
    {InternVL-2.5-78B}~\cite{internvl}  & 31.1 & 38.0 & 31.4 & 21.7 & 21.0 & 27.2 & 40.3  & 29.6 \\
    {Qwen2-VL-72B-Instruct}~\cite{qwen2vl}  & 20.1 & 30.4 & 26.0 & 18.8 & 19.9 & 25.0 & 38.0  & 24.5 \\
    {QvQ-72B-Preview}~\cite{qvq}&  30.1 & 44.6 & 32.1 & 23.9 & 28.0 & 25.0 & 41.1 & 31.0 \\
    {Virgo-72B}~\cite{virgo} & 44.6 & 47.8 & 37.5 & 29.0 & 23.1 & 38.9 & 41.1  & 37.4 \\
   \hline
   \multicolumn{9}{c}{\bf{Open-sourced $\sim$7B MLLMs}} \\
    \hline
    {Qwen2.5-VL-7B-Instruct}~\cite{qwen25vl} & 28.0& 34.8 & 28.6 & 26.1 & 17.2 & 28.9 & 46.5  & 28.8 \\
    {Qwen2-VL-7B-Instruct}~\cite{qwen25vl} & 16.3 & 13.0 & 14.6 & 22.5 & 15.6 & 32.8 & 27.1  & 19.5 \\
    {Qwen2.5-VL-7B-Instruct}+VIC~\cite{vic} & 30.1 & 31.5 & 26.0 & 23.2 & 17.7 & 24.4 & 39.5  & 26.9 \\
    {Qwen2.5-VL-7B-Instruct}+Reflection~\cite{selfcorrection} & 23.9 & 31.5 & 24.8 & 2.3 & 17.2 & 31.7 & 39.5  & 26.7 \\
    {R1-OneVision-7B}~\cite{internvl} & 20.1 & 22.8 & 21.0 & 26.1 & 16.1 & 28.3 & 33.3  & 22.9 \\
    {Mulberry-Qwen2-VL-7B}~\cite{mulberry} &  19.7 & 26.1 & 27.0 & 22.5 & 19.9 & 19.4 & 24.8 & 22.6 \\
    {InternVL-2.5-8B}~\cite{internvl} & 11.4 & 26.1 & 22.5 & 16.7 & 17.2 & 30.6 & 30.2  & 20.8 \\
    {Llama-3.2-Vision-11B}~\cite{llama3}  &  12.5 & 21.7 & 20.0 & 14.5 & 16.1 & 17.2 & 38.0 & 18.7 \\
    {Llava-CoT-11B}~\cite{llavacot}  & 12.1 & 14.1 & 19.7 & 10.9 & 16.1 & 20.0 & 31.0  & 17.4 \\
    {{\method}-7B (Ours)} & 39.1 & 39.1 & 38.7 & 32.6 & 20.4 & 34.4 & 59.7  &  37.1   \\
    \hline
   \multicolumn{9}{c}{\bf{Open-sourced $\sim$3B MLLMs}} \\
    \hline
    {Qwen2.5-VL-3B-Instruct}~\cite{qwen25vl} & 10.7 & 27.2 & 20.6 & 25.4 & 16.1 & 16.1 & 27.1  & 18.8 \\
    {Phi-3.5-Instruct}~\cite{phi3} & 4.1 & 16.3 & 14.0 & 15.2 & 14.5 & 13.9 & 21.7  & 12.9 \\
    {{\method}-3B (Ours)} & 27.3 & 38.0 & 31.4 & 24.6 & 17.7 & 27.2 & 48.0  & 29.4    \\
   \Xhline{1.5pt}
\end{tabular}
}
}
\label{table:per_topic_accuracy}
\end{table}

\subsection{Effect of RL algorithms }As mentioned in Sec.~\ref{sec:method}, the multiple sampling pipeline in the GRPO algorithm is naturally aligned with our hallucination mitigation proposal, \emph{i.e.}, encouraging models to predict along the correct chain for correct answers. To verify the effect of RL algorithms, we compare Logos-7B using GRPO~\cite{deepseekmath} and that using PPO~\cite{ppo}. We remove all CHI stages and directly assess the effect of RL. 
As shown in Table~\ref{table:ablation_rl_algo}, Logos-7B using GRPO surpasses that using PPO by 5.8 and 2.6 on {\dataset} and MathVista, respectively. These results verify our motivation and provide strong support for the Logos framework design. 
We also evaluate the effect of recently proposed DAPO~\cite{dapo}, a specially-designed GRPO variant. Nevertheless, it does not lead to better performance on both benchmarks. A possible explanation is that the newly introduced constraints in DAPO lead to overfitting in the model and restrict the final performance.

\begin{table}[ht]
    \centering
    \caption{LLMs Hallucination Score (LHS) comparison of each dimension in {\dataset}. }
\setlength{\tabcolsep}{3pt} %
    \renewcommand{\arraystretch}{3.5}%
    { \fontsize{8.3}{3}\selectfont{
\begin{tabular}{l|ccccc|c}
   \Xhline{1.5pt}
   {\bf Model}  & {\bf Factual} & {\bf{Logical}} & {\bf{Reasoning}} &  {\bf{Conceptual}} 
    &{\bf Appropriateness}& {\bf Overall}   \\
   \Xhline{1.5pt}
   
   
   \multicolumn{7}{c}{\bf{Black-Box MLLMs}} \\
    \hline
   Gemini-2-Flash-Thinking~\cite{gemini} & 0.7182 & 0.7558 & 0.7689 & 0.7349 & 0.7372 & 0.7517  \\
   O1~\cite{o1} & 0.6054 & 0.6427 & 0.5793 &0.6384 &0.6306 &  0.6193 \\
   Gemini-2-Flash~\cite{gemini} & 0.6640 & 0.7053 & 0.7265 & 0.6862 & 0.7007 &  0.6882  \\
   GPT-4o~\cite{gpt-4o} & 0.5811 & 0.6466 & 0.6777 & 0.6198 & 0.6404 & 0.6332 \\
   \hline
   \multicolumn{7}{c}{\bf{Open-sourced $\sim$72B MLLMs}} \\
    \hline
    {Qwen2.5-VL-72B-Instruct}~\cite{qwen25vl} & 0.6330 & 0.7464 & 0.7912 & 0.6793 & 0.7321 & 0.7233 \\
    {InternVL-2.5-78B}~\cite{internvl} & 0.5830 & 0.6441 & 0.6700 & 0.6088 & 0.6313 & 0.6377 \\
    {Qwen2-VL-72B-Instruct}~\cite{qwen2vl} & 0.4665 & 0.5115 & 0.5339 & 0.4746 & 0.4774  & 0.4928 \\
    {QvQ-72B-Preview}~\cite{qvq} & 0.5024 & 0.5495 & 0.5698 & 0.5368 & 0.5168& 0.5717 \\
    {Virgo-72B}~\cite{virgo} & 0.6094 & 0.6185 & 0.6437 &0.6252 & 0.6187 & 0.6328 \\
   \hline
   \multicolumn{7}{c}{\bf{Open-sourced $\sim$7B MLLMs}} \\
    \hline
    {Qwen2.5-VL-7B-Instruct}~\cite{qwen25vl} & 0.5333 & 0.6201 & 0.6765 & 0.5786 & 0.6130 & 0.5996 \\
    {Qwen2-VL-7B-Instruct}~\cite{qwen25vl} & 0.3512 & 0.3960 & 0.4120 & 0.3573 & 0.3519 & 0.3633 \\
    {Qwen2.5-VL-7B-Instruct}+VIC~\cite{vic} & 0.4600 & 0.4746 & 0.4336 & 0.4449 & 0.4261 & 0.4478 \\
    {Qwen2.5-VL-7B-Instruct}+Reflection~\cite{selfcorrection} & 0.5658 & 0.6242 & 0.6008 & 0.5806 & 0.6117 & 0.5826 \\
    {R1-OneVision-7B}~\cite{internvl} & 0.4565 & 0.5227 & 0.5809 & 0.4822 & 0.5070 & 0.5098 \\
    {Mulberry-Qwen2-VL-7B}~\cite{mulberry} & 0.4545 & 0.4819 & 0.5070 & 0.4605 & 0.4660 & 0.4740 \\
    {InternVL-2.5-8B}~\cite{internvl} & 0.4515 & 0.4967 & 0.5317 & 0.4636 & 0.4757 & 0.4838 \\
    {Llama-3.2-Vision-11B}~\cite{llama3} & 0.4014 & 0.4473 & 0.4741 & 0.4030 & 0.4066 & 0.4265  \\
    {Llava-CoT-11B}~\cite{llavacot} & 0.4050 & 0.4417 & 0.4735 & 0.4116 & 0.4267 & 0.4267 \\
    {{\method}-7B (Ours)} & 0.5841 & 0.6533 &0.7052 & 0.6233 & 0.6566 &  0.6568  \\
    \hline
   \multicolumn{7}{c}{\bf{Open-sourced $\sim$3B MLLMs}} \\
    \hline
    {Qwen2.5-VL-3B-Instruct}~\cite{qwen25vl} & 0.3282 & 0.3593 & 0.3712 & 0.3279 & 0.3242 & 0.3422 \\
    {Phi-3.5-Instruct}~\cite{phi3} & 0.2983 & 0.3443 & 0.3459 & 0.3049 & 0.2968 & 0.3181 \\
    {{\method}-3B (Ours)} & 0.5486 & 0.5947 & 0.6411 & 0.5600 & 0.5757 &  0.5840   \\
   \Xhline{1.5pt}
\end{tabular}
}
}
\label{table:per_dim_lhs}
\end{table}

\subsection{Effect of KL Divergence }
Since we remove the KL-divergence term in {\method} training, to analyze the effect, we conduct an ablation study on the KL-divergence weight. As shown in Table~\ref{table:ablation_KL}, when gradually increasing the weight of the KL-divergence term, the accuracy on both datasets is gradually decreased. When the KL-divergence term is relatively large (\emph{e.g.}, 1e-2), the accuracy on MathVista is even slightly lower than the base model (68.2). A possible explanation is that the distribution of reasoning MLLM has a non-negligible gap with the corresponding base models. To mitigate original reasoning hallucination and bring inherent reasoning capabilities, one should disable the KL-divergence term to tolerate the distribution gap between two models. 

\subsection{Effect of online reward filtration. }
Next, we explore the effect of online reward filtration and report experimental results in Table~\ref{table:ablation_orf}. After integrating ORF into training, {\method}-7B surpasses the counterpart by 2.9 on {\dataset} and 2.7 on MathVista. This improvement proves the effectiveness of ORF and points the future direction for more effective RL algorithms. 

\subsection{Effect of curriculum learning. }
We also conduct experiments to verify the necessity of the CRFT stage. Specifically, we conduct vanilla RL training on all training data and ensure the number of training epochs is equal to the total CRFT. As shown in Table~\ref{table:curriculum_vs_longrl}, even using longer training, one-stage RL still falls behind {\method}-7B with CRFT by 1.6 on {\dataset} accuracy and 0.9 on MathVista accuracy. These results indicate that, benefiting from multi-stage difficulty filtration, the learning efficiency of {\method} is highly improved. 
Meanwhile, we explore the effect of the curriculum learning stage $k$ in {\method}. As shown in Table~\ref{table:ablation_curriculum}, {\method}-7B can easily obtain 37.1 on {\dataset} and 72.3 on MathVista. Further increasing $k$ only introduces a marginal improvement. These results prove the effectiveness and efficiency of CRFT. Therefore, we select $k=1$ to optimize {\method}. 

\subsection{The Quality of Automatic Annotation}
We also assess the accuracy of automatically annotated reasoning chains in different phase, which is shown in Table~\ref{table:annotation_quality}. 
The accuracy of O3-mini initialized reasoning chains achieves 43.9. By incorporating DeepSeek-R1 and guided by answers, the accuracy of refined reasoning chains achieves 73.7.
The relative high accuracy of refined reasoning chains ensures that one can reduce the human labor to correct the reasoning chains with reasoning hallucinations. 

\paragraph{Annotation cost. } Finally we also concern the detailed annotation cost. By using the annotation method proposed in Sec.~\ref{sec:benchmark_annotation}, the total cost is $\sim$22\$. And the total human working hour is 36 hours$*$person. We also estimate the annotation cost using O1, which is nearly 200\$. And if all the questions are annotated by human experts, the total working hour is nearly 200 hours$*$person. These results show the efficiency of our annotation method. 

\begin{table}[h]
   \renewcommand\arraystretch{1.3}
   \centering
   \caption{The reasoning annotation accuracy in each phase.
   }
    \vspace{0.5em}
   \setlength{\tabcolsep}{10.6pt} 
	\renewcommand{\arraystretch}{2.2}
	  		{ \fontsize{8.3}{3}\selectfont{
   \begin{tabular}{l|c}
      \toprule
      \bf {Annotation Phase} & \textbf{Reasoning Chain Accuracy}  \\
      \midrule
      O3-mini (init) & 43.9  \\
       +DeepSeek-R1 (refine w/ answer) & 73.7  \\
      \bottomrule
   \end{tabular}}}
   \label{table:annotation_quality}
\end{table}

\section{Dataset Examples}\label{sec:dataset_example}

\begin{figure}
\begin{center}
\includegraphics[width=\textwidth]{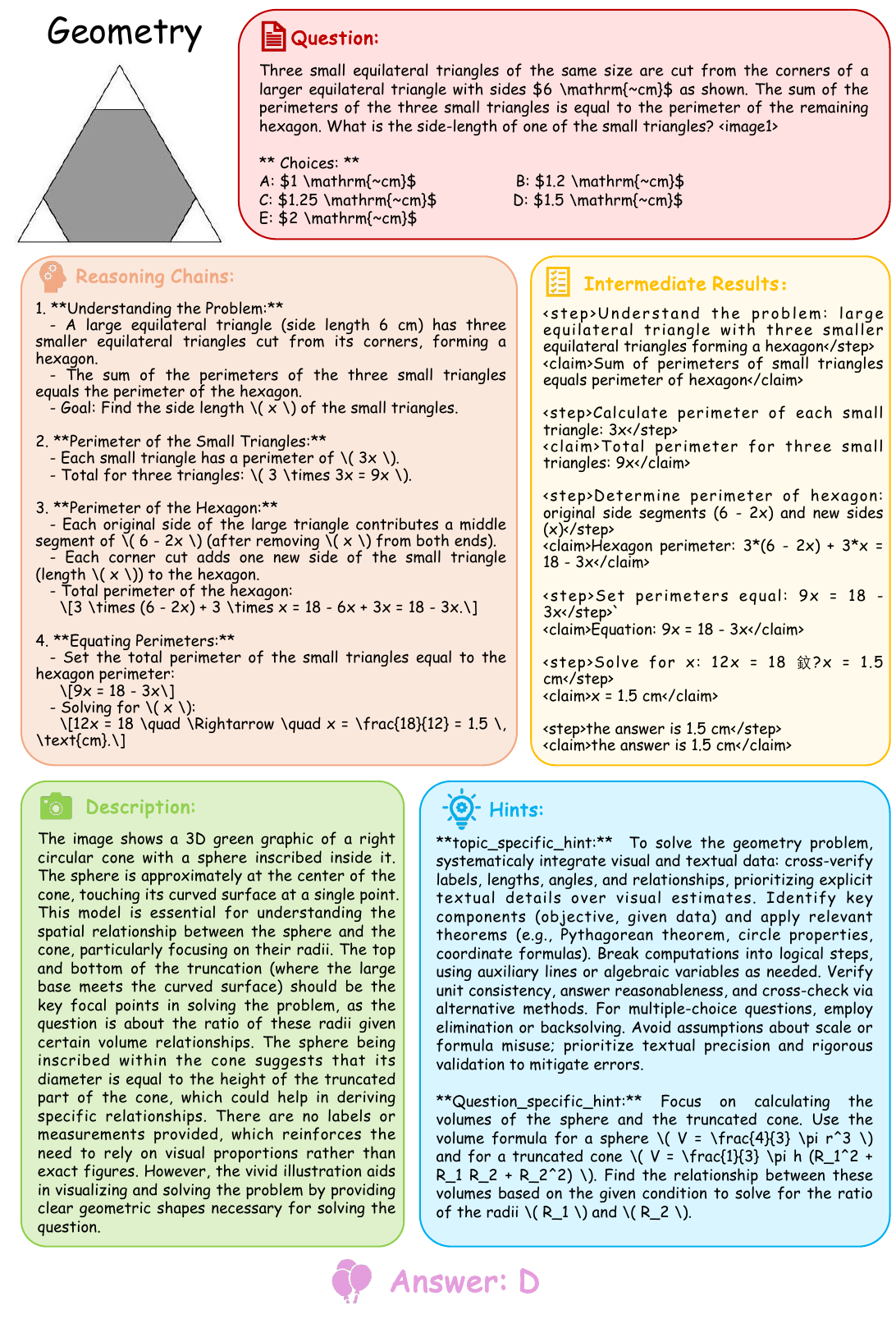}
\end{center}
\caption{The example of geometry question in {\dataset}. 
}
\label{fig:geometry_problem}
\end{figure}

\begin{figure}
\begin{center}
\includegraphics[width=\textwidth]{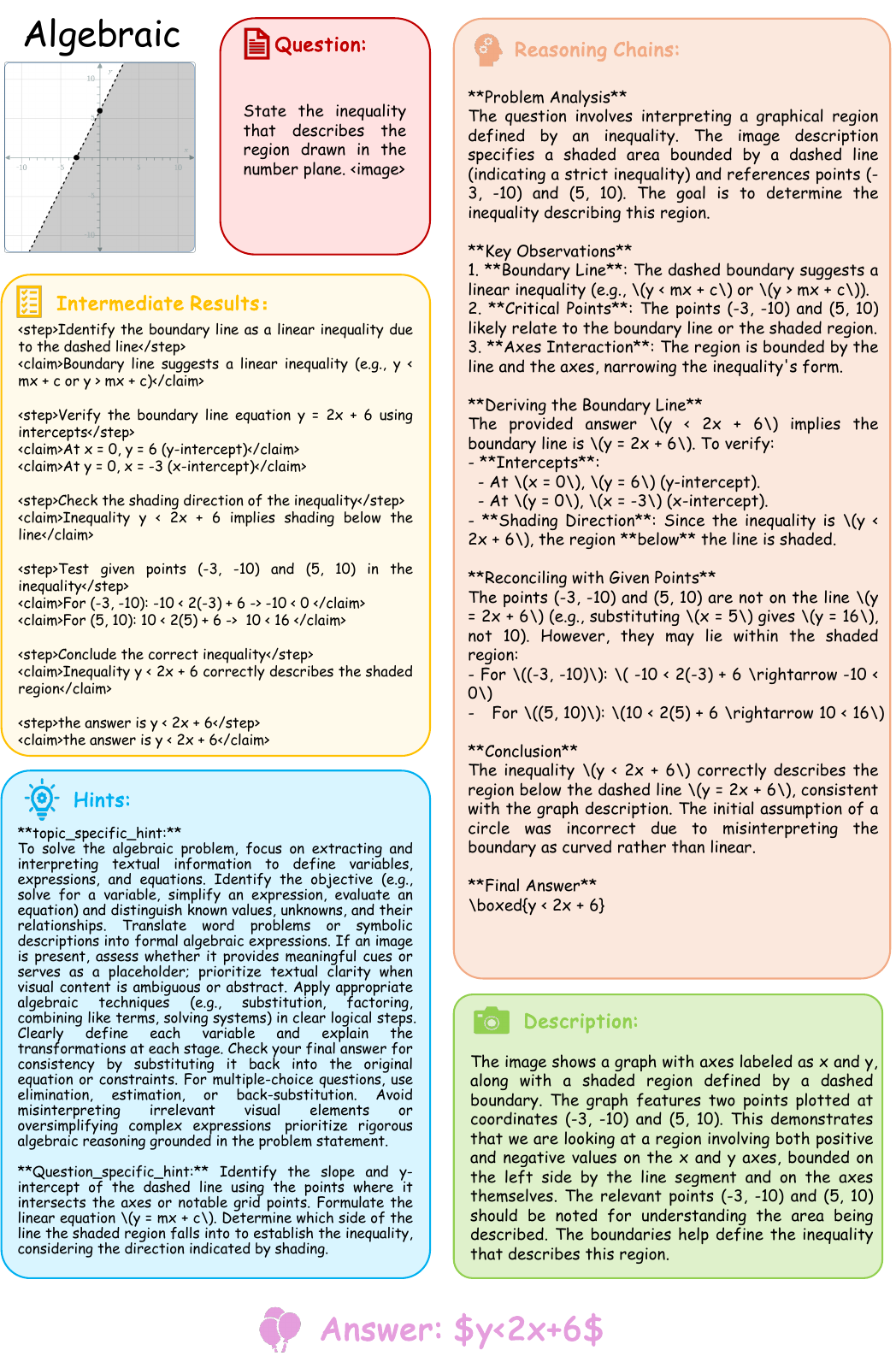}
\end{center}
\caption{The example of algebraic question in {\dataset}. 
}
\label{fig:algebraic_problem}
\end{figure}

\begin{figure}
\begin{center}
\includegraphics[width=\textwidth]{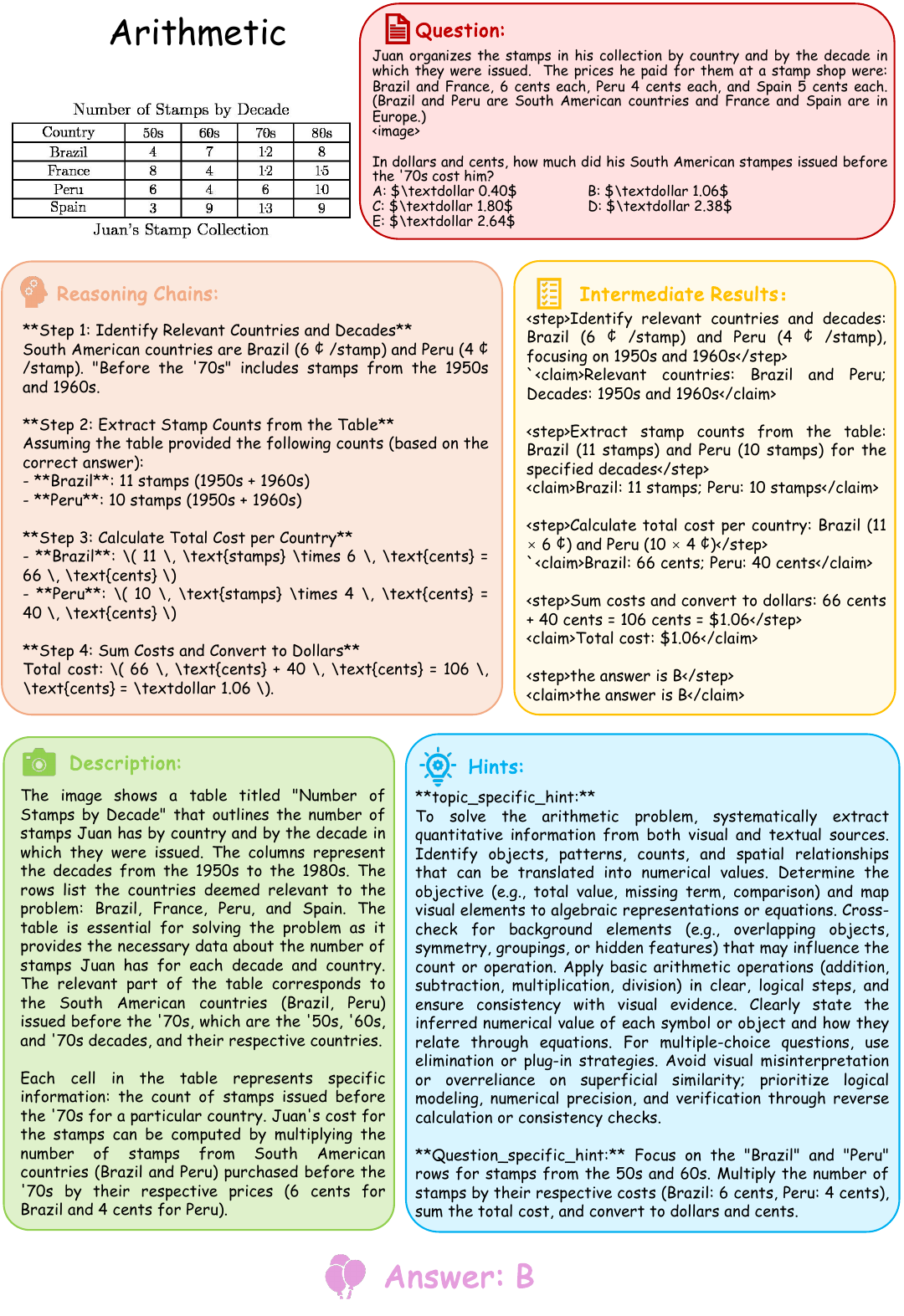}
\end{center}
\caption{The example of arithmetic question in {\dataset}. 
}
\label{fig:arithmetic_problem}
\end{figure}

\begin{figure}
\begin{center}
\includegraphics[width=\textwidth]{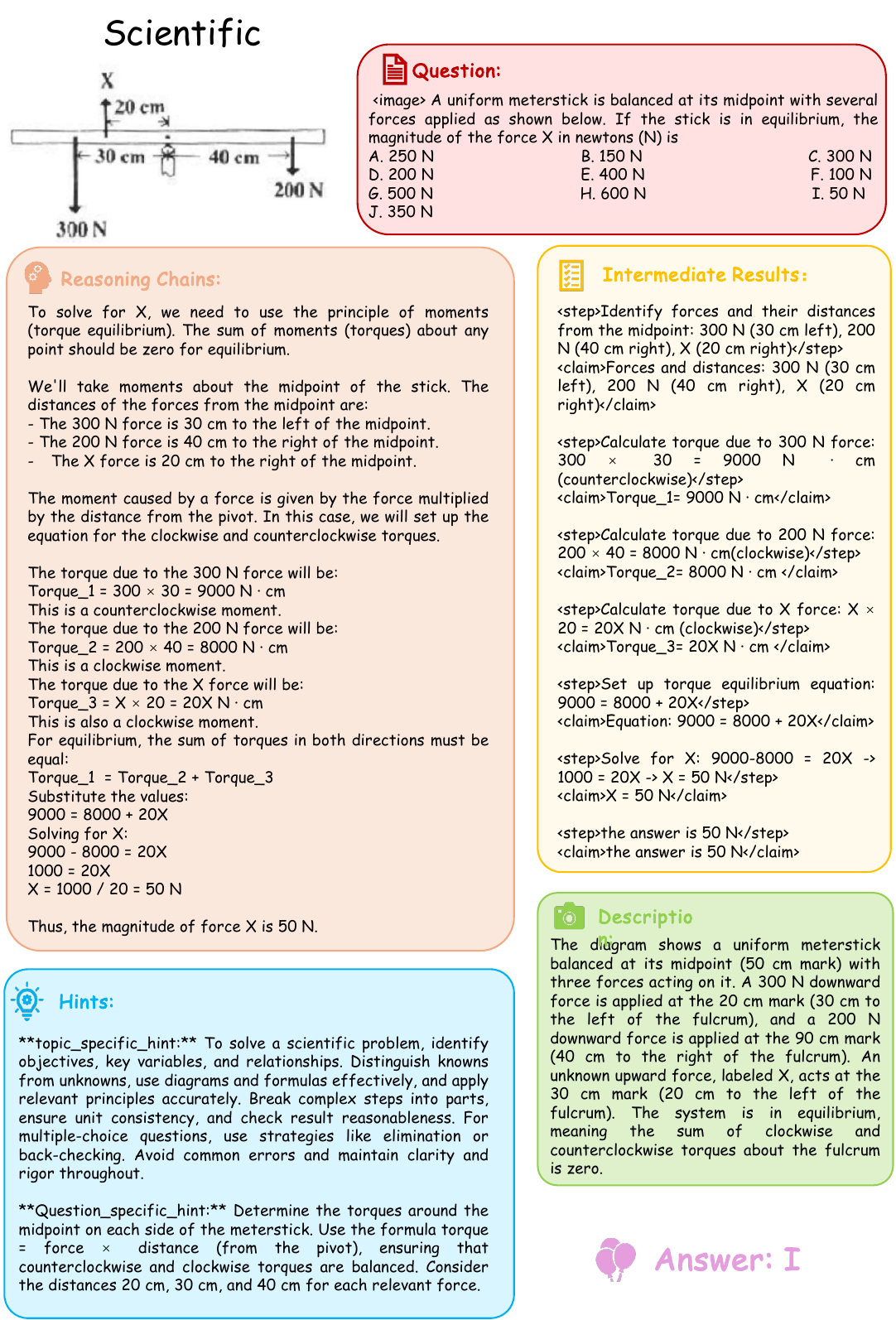}
\end{center}
\caption{The example of scientific question in {\dataset}. 
}
\label{fig:scientific_problem}
\end{figure}

\begin{figure}
\begin{center}
\includegraphics[width=\textwidth]{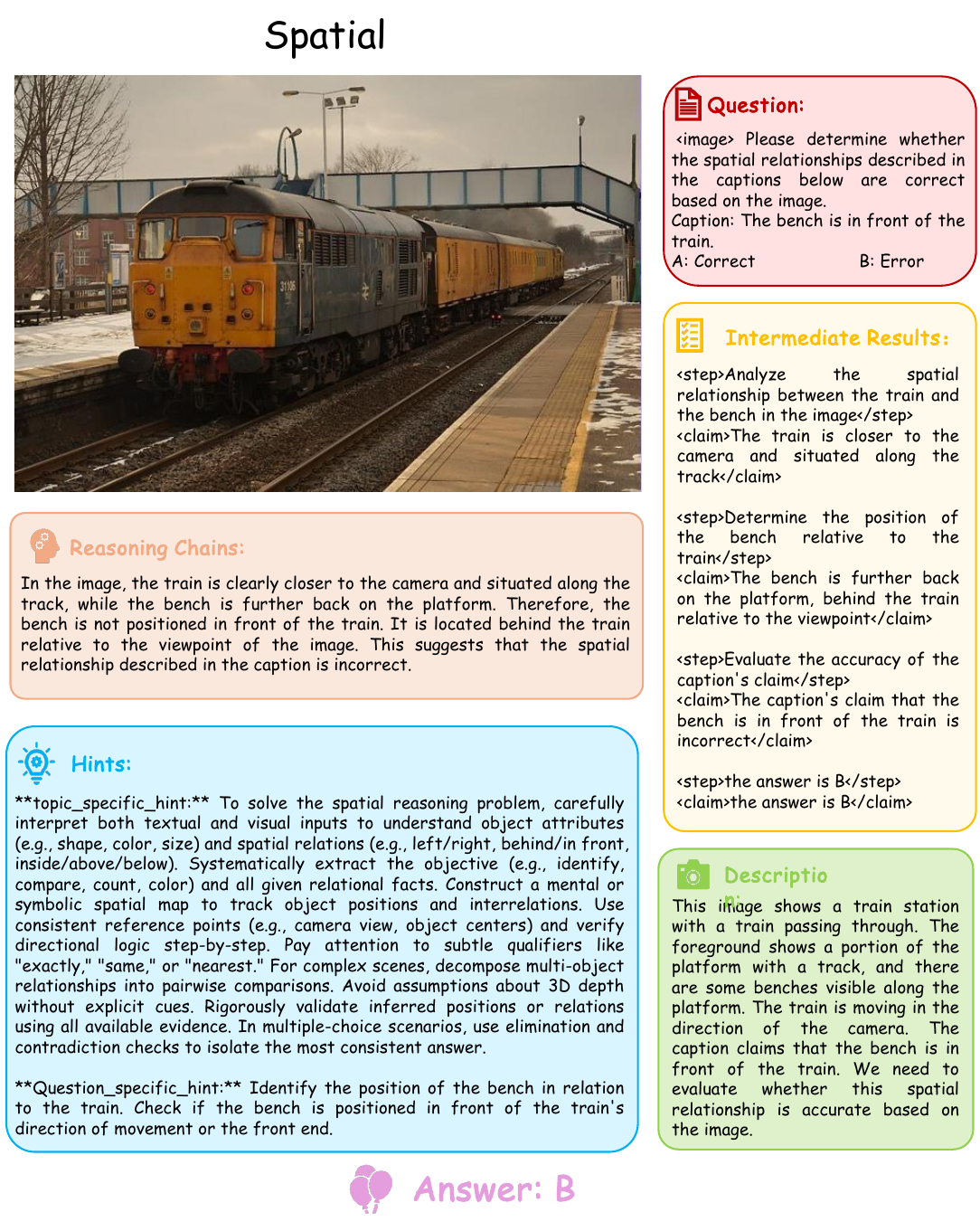}
\end{center}
\caption{The example of spatial question in {\dataset}. 
}
\label{fig:spatial_problem}
\end{figure}

\begin{figure}
\begin{center}
\includegraphics[width=\textwidth]{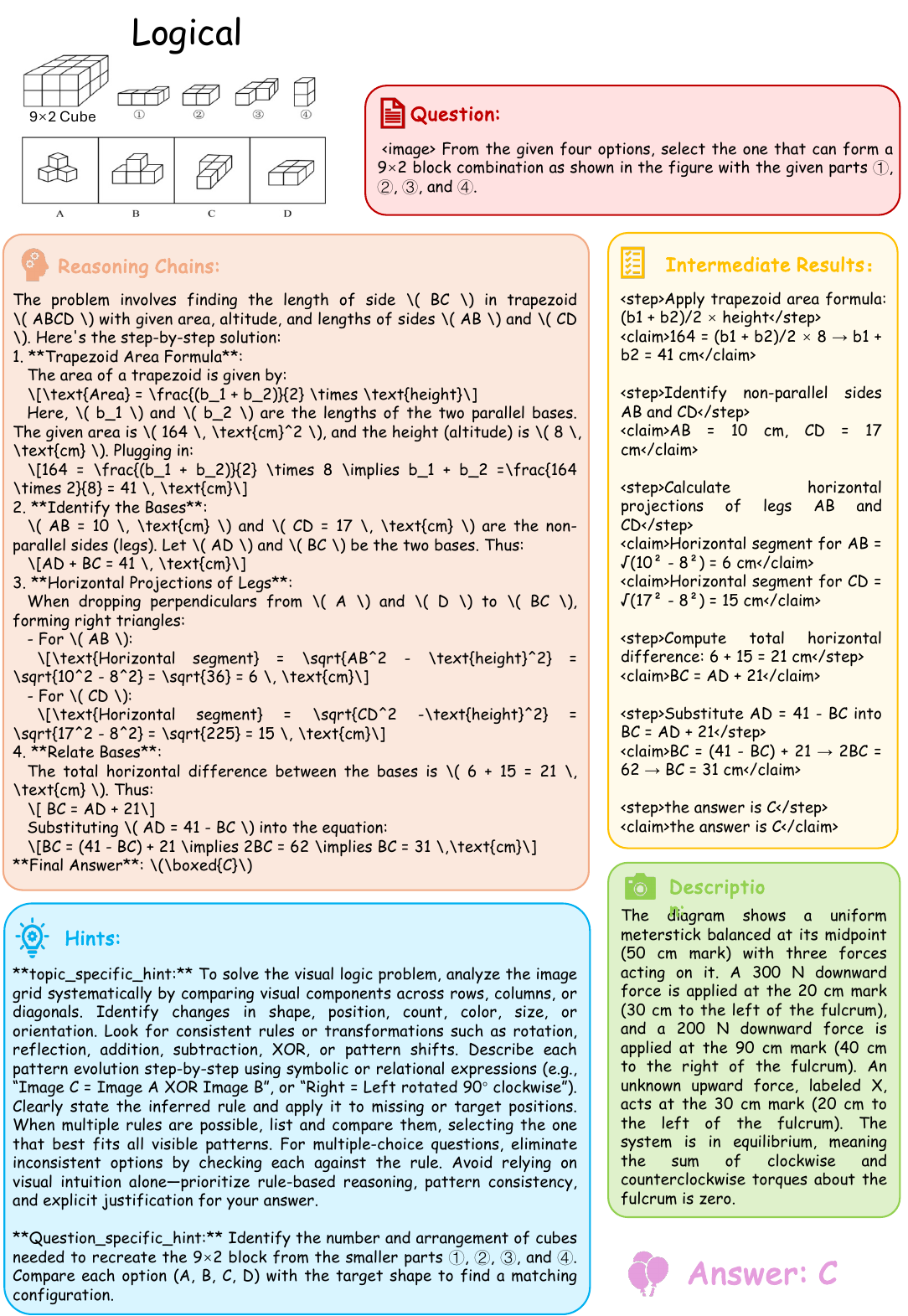}
\end{center}
\caption{The example of logical question in {\dataset}. 
}
\label{fig:logical_problem}
\end{figure}

\begin{figure}
\begin{center}
\includegraphics[width=\textwidth]{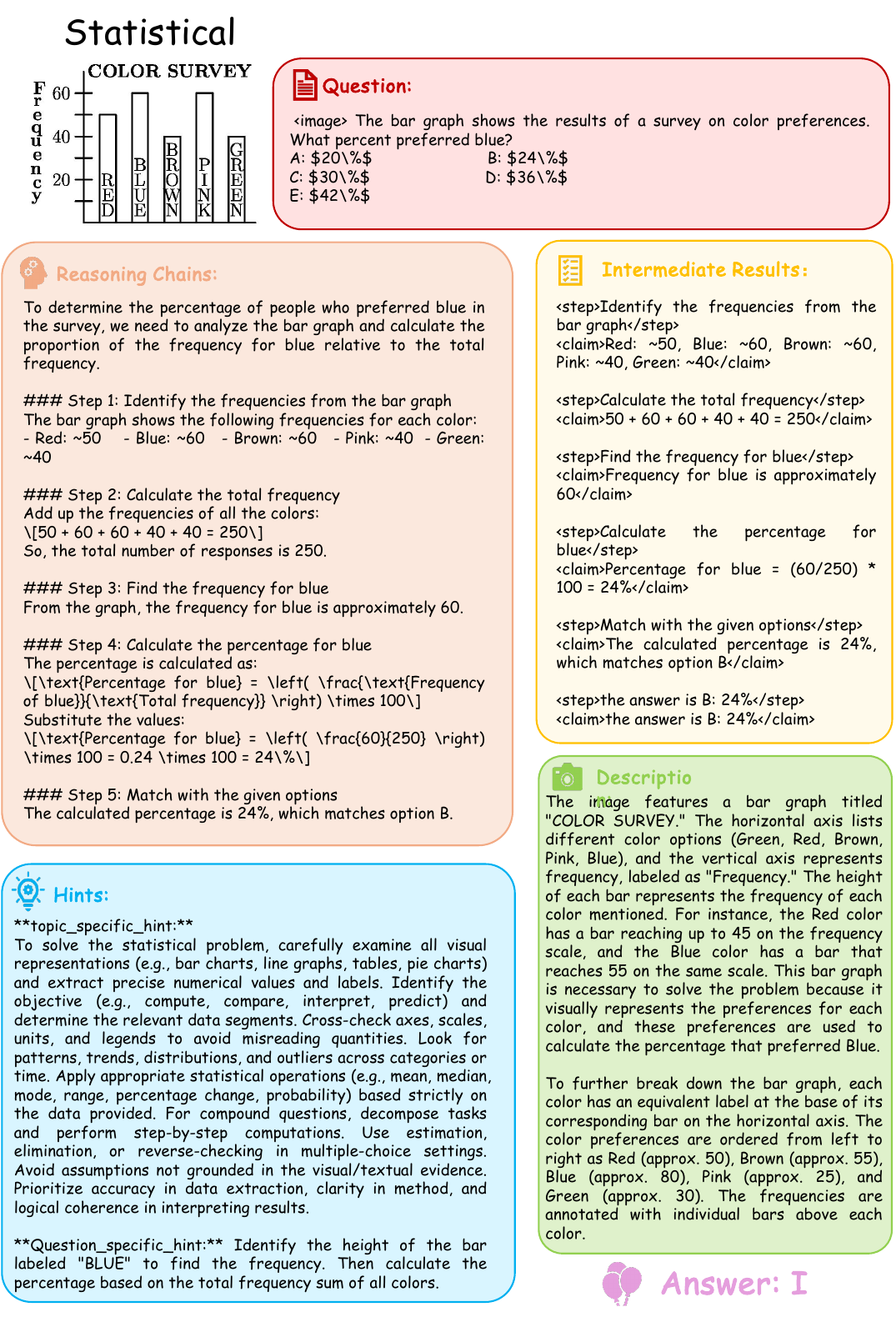}
\end{center}
\caption{The example of statistical question in {\dataset}. 
}
\label{fig:statistical_problem}
\end{figure}

To clearly show the structure of {\dataset}, we provide detailed examples of {\dataset}. Fig.~\ref{fig:geometry_problem} shows an example of geometry questions. 
Fig.~\ref{fig:algebraic_problem} shows an example of algebraic questions. 
Fig.~\ref{fig:arithmetic_problem} shows an example of arithmetic questions. 
Fig.~\ref{fig:scientific_problem} shows an example of scientific questions. 
Fig.~\ref{fig:spatial_problem} shows an example of spatial questions. 
Fig.~\ref{fig:logical_problem} shows an example of logical questions. 
Fig.~\ref{fig:statistical_problem} shows an example of statistical questions. 

\begin{figure}
\begin{center}
\includegraphics[width=\textwidth]{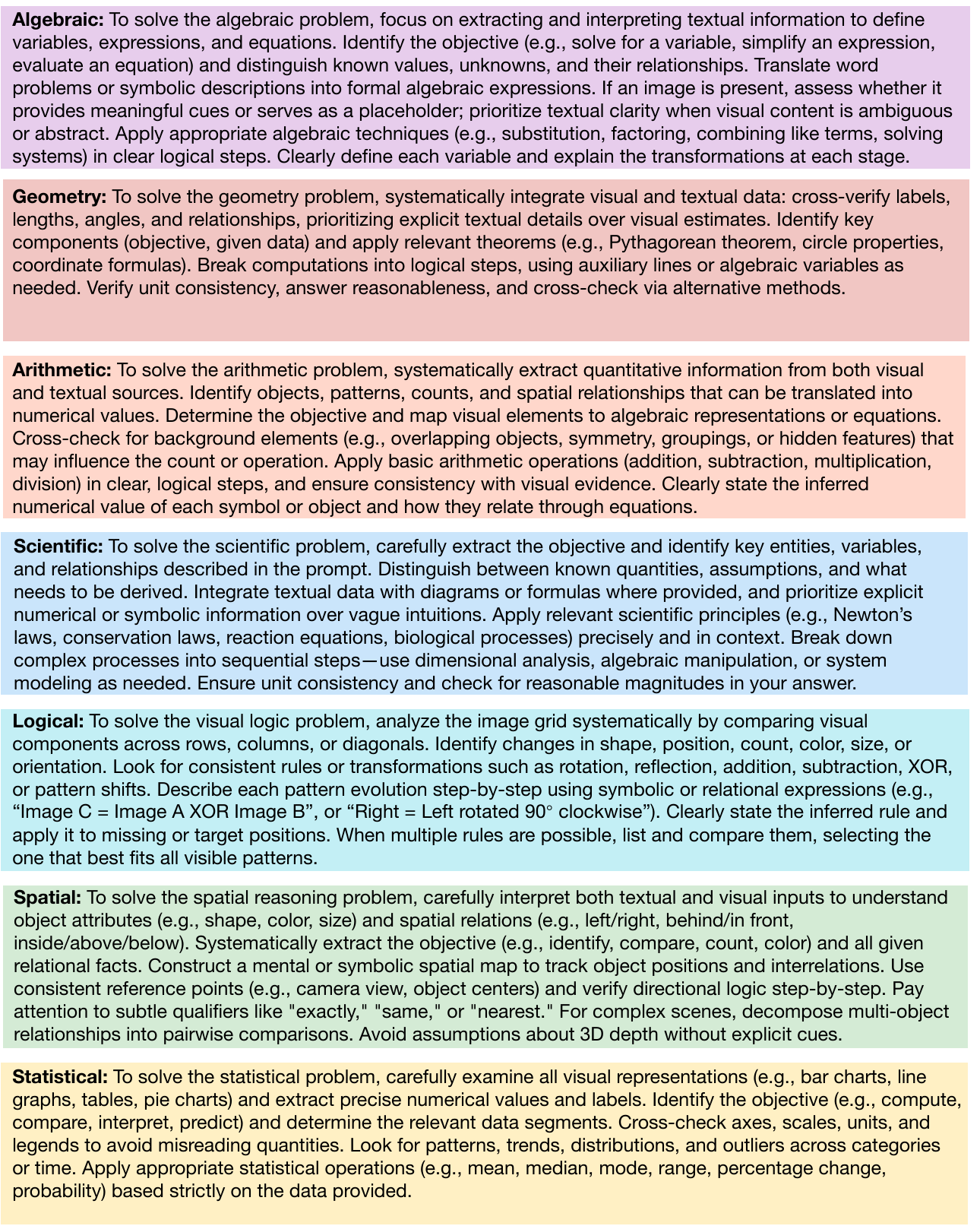}
\end{center}
\caption{The topic-specific hints used in {\dataset}. 
}
\label{fig:topic_hint}
\end{figure}

\paragraph{Summary of Topic-specific Hint. } We also release the topic-specific hints used in {\dataset} and {\method}. As shown in Fig.~\ref{fig:topic_hint}. The topic-specific hints include key concepts and basic rules regarding the question topics. Meanwhile, the classical reasoning process of corresponding question topics is also included in the hints. 

\section{More Qualitative Results}\label{sec:qualitative}
\begin{figure}
\begin{center}
\includegraphics[width=\textwidth]{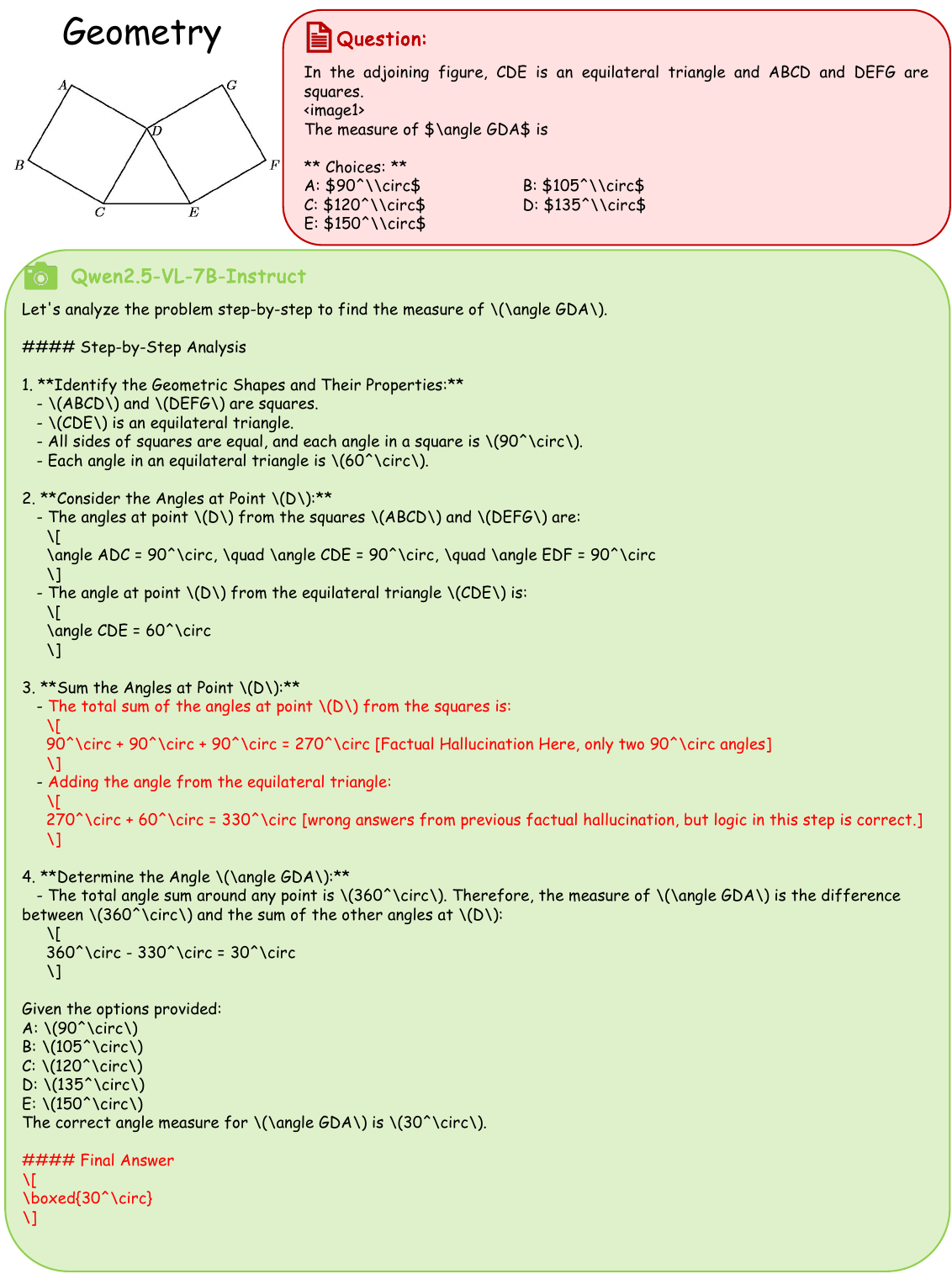}
\end{center}
\caption{Response example from Qwen2.5-VL-7B-Instruct. Red font means reasoning hallucination and corresponding judgement. 
}
\label{fig:qwen_wa_example}
\end{figure}

\begin{figure}
\begin{center}
\includegraphics[width=\textwidth]{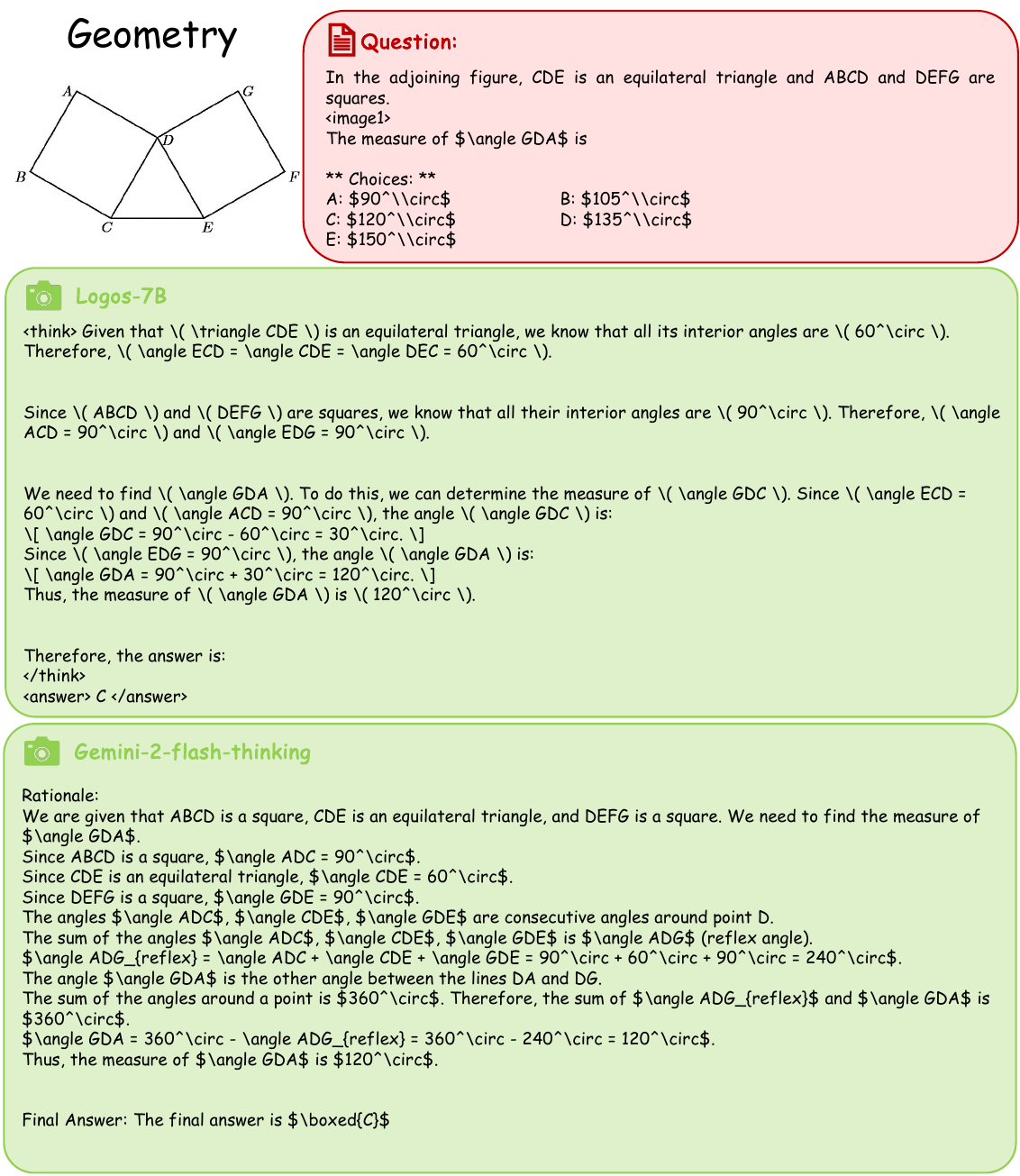}
\end{center}
\caption{Response examples from Logos-7B and Gemini-2-flash-thinking. 
}
\label{fig:logos_gemini_example}
\end{figure}

We also illustrate a couple of raw outputs from some representative models, \emph{i.e.}, Qwen2.5-VL-7B-Instruct, Gemini-2-flash-thinking, and our Logos-7B. As shown in the Fig.~\ref{fig:qwen_wa_example} and \ref{fig:logos_gemini_example}. We find that in this example, Qwen2.5-VL-7B has consistent logic in the reasoning chain, but suffers from factual hallucination (only two 90 degree angles should be calculated). In contrast, Gemini-2-flash-thinking and Logos-7B correctly solve the question. 

\section{Limitation}\label{sec:limitation}
The limitation of this paper is two-fold. First, {\dataset} does not include multiple images or video question-answering problems. Hence, the hallucination from the temporal dimension and the hallucination regarding cross-image relations are not fully explored. And second, the theoretical analysis of why MLLMs suffer from reasoning hallucination is still insufficient. 
These limitations motivate us to conduct more in-depth exploration in the future. 

\section{Broader Impact}\label{sec:impact}
The broader impact of this paper lies in advancing the reliability and accuracy of multimodal large language models (MLLMs) by systematically isolating and evaluating reasoning hallucinations. {\dataset} offers a targeted benchmark for diagnosing and mitigating reasoning errors, which is essential for applications in fields like autonomous systems, medical imaging, and scientific discovery, where accurate multimodal reasoning is critical. By revealing key weaknesses in current MLLMs, such as their struggles with complex spatial reasoning, our work encourages the development of more robust, transparent, and context-aware AI systems, ultimately promoting safer and more trustworthy AI deployment.



